\documentclass[conference,compsoc]{IEEEtran}

\ifCLASSOPTIONcompsoc
  \usepackage[nocompress]{cite}
\else
  \usepackage{cite}
\fi

\usepackage[utf8]{inputenc} %
\usepackage[T1]{fontenc}    %
\usepackage{hyperref}       %
\usepackage{url}            %
\usepackage{booktabs}       %
\usepackage{amsfonts}       %
\usepackage{nicefrac}       %
\usepackage{microtype}      %
\usepackage{xcolor}         %

\usepackage{graphicx}
\usepackage{subcaption}
\usepackage{multirow}

\usepackage{hyperref}

\ifCLASSINFOpdf
\else
\fi

\usepackage{amsmath}
\usepackage{amssymb}
\usepackage{mathtools}
\usepackage{amsthm}
\usepackage{tcolorbox}  

\usepackage[capitalize,noabbrev]{cleveref}

\theoremstyle{plain}

\theoremstyle{definition}

\theoremstyle{remark}

\newtcbox{\redroundedbox}{on line,  
  colframe=red,  
  colback=white,  
  boxrule=1.8pt,  
  arc=4pt,  
  boxsep=0pt,  
  left=2pt,  
  right=2pt,  
  top=2pt,  
  bottom=2pt,  
  before upper={\vphantom{dlg}}  
}  
\usepackage[textsize=tiny]{todonotes}

\newcommand{\system}{Obliviate}
\newcommand{\model}{M}
\newcommand{\txt}{T}
\newcommand{\stride}{s}
\newcommand{\mloss}{\mathcal{L}_{\texttt{maintain}}}
\newcommand{\floss}{\mathcal{L}_{\texttt{forget}}}
\newcommand{\fkl}{\mathcal{L}_{\texttt{dist}}}
\newcommand{\ftarget}{\mathcal{L}_{\texttt{prob}}}
\newcommand{\mypara}[1]{\bigskip\noindent{\bf {#1}.}}

\begin{document}
\title{\system{}: Efficient Unmemorization for Protecting Intellectual Property in Large Language Models}

\author{
\IEEEauthorblockN{Mark Russinovich}
\IEEEauthorblockA{Microsoft Azure\\
Mark.Russinovich@microsoft.com}
\and
\IEEEauthorblockN{Ahmed Salem}
\IEEEauthorblockA{Microsoft\\
Ahmsalem@microsoft.com}
}

\maketitle

\begin{abstract}
Recent copyright agreements between AI companies and content creators underscore the need for fine-grained control over language models’ ability to reproduce copyrighted text. Existing defenses—ranging from aggressive unlearning to simplistic output filters—either sacrifice model utility or inadequately address verbatim leakage. We introduce \system{}, a lightweight post-training method that surgically suppresses exact reproduction of specified sequences while preserving semantic understanding.
\system{} first identifies memorized passages and then, for each target token, minimally adjusts the model’s output distribution via a Kullback–Leibler divergence penalty to drive down the probability of exact reproduction. Simultaneously, we enforce a consistency loss on non-target tokens to retain the model’s fluency and task performance. We evaluate \system{} on four popular 6–8B-parameter models (LLaMA-3.1, LLaMA-3.1-Instruct, Qwen-2.5, and Yi-1.5) using synthetic memorization benchmarks and organic copyrighted excerpts (e.g., Moby Dick, Frankenstein, Alice in Wonderland and Les Miserables). Across all settings, \system{} reduces verbatim recall by two orders of magnitude (e.g., from hundreds of words to fewer than 12) while degrading downstream accuracy by at most 1\% on HellaSwag, MMLU, TruthfulQA, and Winogrande. Furthermore, we benchmark \system{} aganist different unlearning and copyright techniques using the MUSE and CoTaEval benchmarks. These results position \system{} as a practical, high-fidelity solution for copyright compliance in deployed LLMs.
\end{abstract}\IEEEpeerreviewmaketitle

\section{Introduction}
\label{sec:intro}

The rise of large language models (LLMs) has intensified debates over intellectual property and security in AI‐generated text. Federal inquiries by the U.S. Copyright Office and the European Union’s proposed AI Act~\cite{USCO2025,EUAIAct2024} highlight the need to distinguish lawful derivative use from infringing reproduction. At the heart of this discussion lies a crucial distinction between derivative works and infringing reproductions when models emit verbatim sequences from their training data~\cite{Stanford2023}. Moreover, adversarial extraction attacks reveal that training data generation not only risks copyright violation but can also leak private or sensitive information~\cite{CTWJHLRBSEOR21}.

Recent legal frameworks and industry accords have begun to address this gap by targeting verbatim replication policies~\cite{plagiarismtoday}. For example, publishers are negotiating acceptable excerpt lengths~\cite{ingramspark,janefriedman}, while leading AI vendors impose strict caps on exact text regeneration, either by blocking copyrighted passages outright or halting generation once a specified word count is reached.

Historically, safeguarding intellectual property in LLMs has relied on post‐processing filters, alignment procedures, or unlearning techniques designed to remove copyrighted/memorized text. In contrast, emerging agreements emphasize protecting specific expressions, e.g., exact generation, rather than abstract ideas. This shift reframes the technical challenge: rather than preventing models from internalizing concepts, we must disable them from generating copyrighted/protected token sequences.

\subsection{Contribution}

In this paper, we introduce \system{}, the first --to the best of our knowledge-- post‐training unmemorization method that selectively suppresses exact token sequences in LLMs while preserving their utility. While recent work —such as the GoldFish loss—has demonstrated promise in suppressing memorization during training\cite{HWJKKSSSGBG24}, \system{} addresses the more common scenario of already‐deployed models. This allows for a quick patching tool to address copyright issues—or memorization problems in general—after deployment.

Intuitively, \system{} makes the LLM ``forget'' multiple tokens within a target span—such as copyrighted content—so that generation inevitably diverges and the probability of reproducing the original text approaches zero. To achieve this, \system{} first selects the target tokens using a fixed-stride sliding window. A larger window size results in changing fewer tokens, thereby better preserving the model's utility, while smaller sizes lead to forgetting more tokens, thus enhancing unmemorization performance. For this work, we find that a window size of four tokens achieves an optimal balance between these two objectives. Nonetheless, the window size is a hyperparameter that can be tuned for different applications.

After selecting the target tokens, we introduce a novel loss function called the ``forget'' loss ($\floss$). For each flagged token, we remove its highest-probability entry from the model’s output distribution and then minimize the Kullback–Leibler divergence between the original top-(k) distribution and the modified one ($\fkl$). We selectively choose the candidate tokens for these target tokens to ensure that certain properties, such as capitalization, are preserved. An additional term is incorporated to penalize any residual probability mass on the target token, further driving its likelihood toward zero, namely the probability loss ($\ftarget$), i.e.,
\[
\floss = \fkl + \ftarget
\]

To maintain the overall model behavior, we concurrently apply a ``maintain'' loss, \(\mathcal{L}_{\text{maintain}}\), which is also based on KL divergence but focuses on non-target tokens. Optimizing the combined objective
\[
\mathcal{L} = \mathcal{L}_{\text{forget}} + \mathcal{L}_{\text{maintain}}
\]
enables selective suppression of verbatim replication while retaining fluency and factual knowledge.

We extensively evaluate \system{} on both synthetic and organic datasets across multiple architectures and scales, including LLaMA‐3.1 8B, LLaMA‐3.1‐Instruct 8B, Qwen‐2.5 7B, and Yi‐1.5 6B. Our evaluation demonstrates that \system{} significantly reduces the generation probability of copyrighted content, with the longest common string length dropping from 270-380 to under 12 for syntactic data. We observe similarly strong performance on organic content, such as naturally occurring copyrighted texts (e.g., \textit{Frankenstein}, \textit{Moby Dick}, \textit{Alice in Wonderland}, and \textit{Les Miserables}). Furthermore, we assess the impact of \system{} on the utility of the models across standard benchmarks (MMLU, TruthfulQA, HellaSwag, Winogrande). Our results show that \system{} maintains the utility of the models, with no drop in most cases and a maximum drop of around 1\%. Finally, we benchmark \system{} using various metrics such as CoTaEval \cite{wei2024CoTaEval} and MUSE \cite{shi2024muse}, which compares different techniques for enforcing copyright protection. The results shows that \system{} achives the best utility-unmemorization trade off across all different techniques, e.g., Gradient Ascent (Grad Ascent)\cite{Thudi22gradAscent}, Gradient Difference (Grad Diff)\cite{liu202graddiff}, KL Minimization (KL)\cite{golatkar2020KL}, Preference Optimization (PO)\cite{rafailov2024idk}, negative preference optimization (NPO)~\cite{zhang2024NPO}, and SimNPO~\cite{fan2025SimNPO}.

In summary, our contributions are:
\begin{itemize}
    \item We propose \system{}, a novel post‐training unmemorization framework that targets verbatim passages without harming LLM utility.
    \item We demonstrate \system{}'s efficacy and scalability through extensive experiments on multiple LLM families and both synthetic and organic datasets.
\end{itemize}

\subsection{Organization} 
This paper is organized as follows: First, we present the background and preliminaries (\autoref{sec:prelim}), where we also introduce our threat model. Then, we present the related work (\autoref{sec:related}), methodology (\autoref{sec:meth}), and evaluation (\autoref{sec:eval}) sections. Finally, we discuss \system{}'s limitations in \autoref{sec:limitations} and conclude in \autoref{sec:conc}.

\section{Background and Preliminaries}
\label{sec:prelim}
\subsection{Types of Model Memorization}
LLMs exhibit two fundamentally different patterns of memorization during training:

\mypara{Semantic Memorization}  
Semantic memorization occurs when a model memorizes abstract concepts, relationships, or factual knowledge from its training corpus. This behavior is generally beneficial: it allows the model to draw on learned representations to perform reasoning, answer questions, and transfer knowledge to novel contexts.

\mypara{Verbatim Memorization}  
By contrast, verbatim memorization refers to the model’s ability to reproduce exact fragments of text seen during training. While semantic memorization leads to useful generalization, verbatim memorization raises serious concerns for privacy, copyright, and data leakage. It is precisely this capacity to regurgitate training sequences, rather than to generalize over them, that we address in this work.

We formalize verbatim memorization in terms of a model’s ability to complete text sequences given only partial context.  Let $\mathbf{s} = (s_1, \dots, s_n)$ be any token sequence drawn from the training data.  We say $\mathbf{s}$ is \emph{$(\ell,\beta)$--memorized} if there exists a prefix length $\ell$ such that:
\[
P_\model\bigl(s_{i+\ell}, s_{i+\ell+1}, \dots, s_n \mid s_{i:i+\ell-1}\bigr) \;>\; \beta,
\]
where the prefix start index $i$ is sampled uniformly from $\{1,\dots,n-\ell\}$ and $\beta \in [0,1]$ is a probability threshold.

Following the approach of \cite{CTWJHLRBSEOR21}, we further simplify this definition by focusing on greedy decoding.  Given a prefix $s_{i:i+\ell-1}$, let the model generate tokens one at a time by selecting
\[
s_j^{\mathrm{gen}} = \arg\max_{x \in \mathcal{V}} P_\model\bigl(x \mid s_{i:j-1}\bigr)
\quad\text{for }j = i+\ell,i+\ell+1,\dots
\]
and let $n^* \;=\; \max\bigl\{\,n : s_{i:j}^{\mathrm{gen}} = s_{i:j}\ \forall\,j \le i+n\,\bigr\}$ denote the length of exact reproduction.  To capture generation under typical settings, we also evaluate $n^*$ when sampling with the model’s default temperature, rather than strictly greedy decoding.

\subsection{Threat Model}

Our goal is to safeguard against inadvertent disclosure of copyrighted material in typical, non-adversarial settings.  Under this threat model, we seek to minimize the model’s tendency to reproduce protected text. Although a determined attacker might still be able to coerce the model into producing such content, our primary objective is to prevent routine copyright infringements.

\section{Related Work}
\label{sec:related}
Our work builds upon and extends several lines of research in machine learning memorization, unlearning, and copyright protection. We organize related work into three main categories: (1) memorization in language models, (2) machine unlearning approaches, and (3) copyright protection techniques.

\subsection{Memorization in Language Models}
Understanding and measuring memorization in language models has been an active area of research. \cite{CIJLTZ23,ZILJTC23} demonstrated that large language models can memorize and reproduce substantial portions of their training data. \cite{KWR22} and \cite{LINZECC22} showed that memorization is particularly prevalent in commonly used web-scraped datasets, with exact duplicates in training data significantly increasing the likelihood of memorization. These findings raise significant concerns about privacy and copyright protection in deployed language models.

\subsection{Machine Unlearning}
Machine unlearning approaches focus on removing specific data points or concepts from trained models, targeting both semantic and verbatim memorization. These approaches can be categorized into exact and approximate unlearning methods.
Exact unlearning methods provide theoretical guarantees for complete removal of target sequences. \cite{BCCJTZLP20} proposed an efficient retraining approach using data sharding. However, subsequent work\cite{CZWBHZ21} demonstrated that such methods remain vulnerable to membership inference attacks \cite{SSSS17,SZHBFB19} when adversaries have access to both the original and unlearned models.
Approximate unlearning methods trade theoretical guarantees for computational efficiency. \cite{GNG21} introduced Amnesiac Unlearning, which tracks model updates and their corresponding training batches, enabling selective removal by ignoring specific updates. Various other approximate techniques have been proposed \cite{GGHM23,GAS20,WHS22,MPSR22,ER23}, each offering different trade-offs between unlearning effectiveness and computational cost. For this work, we compare against multiple unlearning techniques that are part of the MUSE \cite{shi2024muse} benchmark and the copyright protection benchmark CoTaEval \cite{wei2024CoTaEval}, namely Gradient Ascent (Grad Ascent) \cite{Thudi22gradAscent}, Gradient Difference (Grad Diff) \cite{liu202graddiff}, KL Minimization (KL) \cite{golatkar2020KL}, Preference Optimization (PO) \cite{rafailov2024idk}, Negative Preference Optimization (NPO) \cite{zhang2024NPO}, and SimNPO \cite{fan2025SimNPO}.

While the recent GoldFish loss \cite{HWJKKSSSGBG24} similarly focuses on verbatim memorization, it operates during the training phase and cannot be directly applied to pretrained models. Our work adapts and extends these concepts for post-training modification, introducing novel techniques for maintaining model utility while targeting specific memorized sequences.

\subsection{Copyright Protection}
Recent work on copyright protection in language models has explored several approaches. Watermarking techniques \cite{LWWG23,SDSJ20,SFDDF24} enable detection of copyrighted content by embedding identifiable markers in training data—if a model generates text containing these watermarks, it indicates training on copyrighted content. However, these approaches focus on detection rather than prevention of copyright violations.
In practice, companies primarily rely on filtering mechanisms to detect and block the generation of copyrighted content, such as Azure's content filters, and alignment techniques that teach models to not generate copyrighted content. \system{} improves upon these approaches by modifying the model's parameters directly with a targeted loss function, making it inherently resistant to generating copyrighted content during normal usage without requiring additional computational overhead during inference from classifiers or blocklists.

We believe our work lies between complete unlearning and simple output filtering. While unlearning approaches aim for complete removal of concepts at significant computational cost, and filtering approaches operate only at inference time, \system{} provides an efficient method to modify pretrained models to comply with copyright restrictions without sacrificing utility or requiring expensive retraining. This makes it suitable for patching models to prevent generation of newly identified or reported copyright concerns in deployed models.\begin{figure*}[!t]
\centering
\includegraphics[width=1.5\columnwidth]{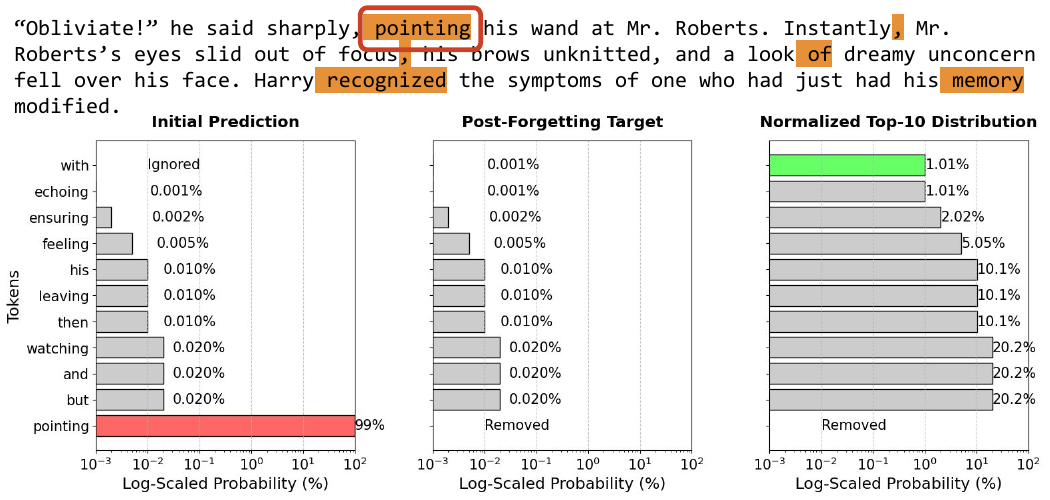}
\caption{
Visualization of \system{}. Top of the figure shows an excerpt from "Harry Potter and the Goblet of Fire," where tokens designated for unmemorization are {\colorbox{orange}{highlighted}}, using a stride of 10. Below, the three-stage process of distribution transformation are shown: (a) the initial prediction distribution($\mathbf{z}$), representing the model's original probabilities for the \redroundedbox{pointing} token; (b) the target distribution after removing the target token, which serves as the post-forgetting target (${\mathbf{z}{\setminus x_\texttt{target}}}$); and (c) the final normalized distribution ($\text{softmax}\left({\mathbf{z}{\setminus x_\texttt{target}}}\right)$) used for computing the forgetting loss $\floss$.
}
\label{fig:mainFig}
\end{figure*}

\section{Methodology}
\label{sec:meth}
\subsection{Problem Formulation}
Let $\model$ be a pretrained language model with parameters $\theta$, which may have memorized sensitive or copyrighted sequences $\txt$ from its training data. Our objective is to modify $\model$ so that it no longer reproduces these sequences verbatim, while preserving its ability to generate contextually appropriate text—i.e., without degrading its utility.

This goal differs from classical unlearning, where the model is expected to completely forget the concept of $\txt$ and not only avoid verbatim reconstruction. Instead, \system{} retains the underlying concept (for example, understanding what “Harry Potter” is and answering related questions) but prevents exact reproduction of text from the Harry Potter books.

We believe that this notion of \emph{unmemorization} simplifies the unlearning problem and aligns with industry practice. For instance, Anthropic has discussed agreements with copyright holders to block exact reconstruction of their content \cite{reuters}.

\subsection{\system{}}

Building on the “GoldFish” loss \cite{HWJKKSSSGBG24}, we adapt it to a post-training setting, where the model already exhibits memorized, copyrighted outputs. Intuitively, \system{} perturbs selected tokens in the memorized text $\txt$ to divert $\model$ away from generating $\txt$ exactly, causing it to produce alternative text.

\mypara{Stride and Token Selection}
First, we define a stride $\stride$ that sweeps over the memorized text $\txt$ in fixed-size windows. This stride determines which tokens will be targeted for unmemorization essentially skipping $\stride$ tokens after each target token to be unmemorized. For each selected token, we apply a \emph{forget loss} $\floss$ to reduce its likelihood, while concurrently using a \emph{maintain loss} $\mloss$ to conserve the model’s overall behavior. The top of \autoref{fig:mainFig} shows an example from a ``Harry Potter'' excerpt with tokens chosen using $\stride=10$.

To ensure that the model's utility remains uncompromised, we selectively retain certain tokens even if they have been identified for unmemorization. Specifically, we exclude special tokens such as ``start of text'' and ``end of text'' to prevent damage to the model's templates. Furthermore, we also exclude tokens with a 100\% probability, as these tokens lack suitable alternative candidates. In other words, any replacement token at this location would be equally probable with a probability of 0\%, rendering the candidate selection random and potentially pushing the model towards an undesired random direction, thus affecting its utility.
\subsubsection{Forget Loss}
\label{sec:forgetLoss}
$\floss$ consists of two different components: the first ($\fkl$) penalizes the output distribution of each target token, while the second ($\ftarget$) penalize the target tokens probability to further decrease them:

\[
\floss = \fkl + \ftarget
\]

\mypara{$\fkl$ (Distribution Loss)} Intuitively, the distribution loss use KL-divergence (KL-div) on the output distribution of the target model, but with the top predicted token (the target token to unmemorize) removed. For this work, we select the top-10 probabilities to apply the KL-div and normalize them to compensate for the effect of removing the top token. 
The bottom of \autoref{fig:mainFig} shows an example of the constructed distributions of a target token `` pointing''.
More formally, $\floss$ is defined as follows:
\[
\fkl = \operatorname{KL}\Bigl(
\operatorname{softmax}(\mathbf{z})
\;\Big\|\;
\operatorname{softmax}(\mathbf{z}\setminus x_{\mathrm{target}})
\Bigr),
\]
where $\mathbf{z}$ represents the top-10 logits of the target model, $\mathbf{z}{\setminus x_\texttt{target}}$ represents the logits after removing the target token ($x_\texttt{target}$), and $x_\texttt{target}$ is the target token to unmemorize.

To select candidates for the $\floss$ loss, we inspect and filter the top-$k$ candidate tokens for each token targeted for unmemorization to ensure they are valid substitutions, i.e., to maintain the model's utility, before applying the forget loss. As we iterate through the highest‐probability tokens (excluding the original target token), we discard any candidates that meet one or more of the following criteria:
\begin{enumerate}
    \item They are special tokens (e.g., end‐of‐text markers).
    \item They violate formatting constraints: if the original token required a leading space or hyphen, the candidate must exhibit the same formatting.
    \item They fail a capitalization check, ensuring that the replacement matches the original token’s casing pattern.
\end{enumerate}
If filtering reduces the number of valid tokens below $k$, we extend our search to the next highest‐probability tokens (i.e., from $k+1$, $k+2$, \dots) until we recover a full set of $k$ candidates for the forget loss.```

\mypara{$\ftarget$ (Probability Loss)}
The second part of the forget loss, the probability loss ($\ftarget$),  aims to further suppress the probability of each target token $p_{i}$. This results in the model being less probable to generate the target sequence/text. Concretely, the target loss ($\ftarget$) is defined as:
\[
\ftarget
= \frac{1}{N}\sum_{i=1}^{N}
\Bigl[
  \lambda_{1}\,\bigl(\log(1 - p_{i} + \varepsilon)\bigr)^{2}
  \;+\;
  \lambda_{2}\,\log\!\bigl(\tfrac{p_{i}}{\tau} + \varepsilon\bigr)\,\tfrac{p_{i}}{\tau}
\Bigr]
\]
where $\lambda_{1},\lambda_{2}$ weight the two terms, $\tau$ is a threshold ($10^{-4}$ in this paper), and $\varepsilon$ is a small constant for numerical stability. Intuitively, the first term pushes to reduce the probability to 0, while the second one add additional penalization to the probability if it is above specific threshold ($\tau$). The second term is set to zero when $p_{i}\le\tau$.

\mypara{Loss Scheduling}
Naively combining the distribution ($\fkl$) and probability ($\ftarget$) results in strong unmemorization of the target tokens, however as we saw in our experiments it degrades the utility of the models. After experimenting with different combination strategies, we found the best one is to first apply the distribution loss on its own until the model adequately forgets the target tokens. Then we combine both losses, i.e:
\[
\floss = \fkl + \ftarget
\]
This way the model first forgets the target tokens while keeping its distribution mostly the same, then further push the target tokens in lower ranks without affecting the utility much (since $\fkl$ already pushed the target tokens low enough and hence $\ftarget$ does not need to push for changing the model's output distribution significantly). In short, this schedule ensures effective unmemorization while minimizing impact on the model’s utility.
\subsubsection{Maintain Loss}
The maintain loss $\mloss$ preserves the model’s overall behavior by minimizing the KL-divergence between the original and updated output distributions over non-target tokens. We favor KL-divergence over cross-entropy to avoid inadvertently reinforcing memorization of non-target tokens.

\bigskip

By balancing these objectives, \system{} unmemorizes selected tokens without compromising model utility. The final objective is
\[
\mathcal{L} = \lambda_f \floss + \lambda_m \mloss
\]
where $\lambda_{f}$ and $\lambda_{m}$ control the trade-off between forgetting and maintenance.

\section{Evaluation}
\label{sec:eval}
\subsection{Experimental Setup}
\subsubsection{Models}
We evaluate \system{} on four state-of-the-art large language models (LLMs) that vary in architecture, scale, and training methodology, providing a comprehensive assessment of \system{}'s robustness across diverse model configurations.

\mypara{LLaMA-3.1 8B (Base (Llama) and Instruction-Tuned (Llama-Inst))\cite{llama3}} Developed by Meta AI, LLaMA-3.1 8B is an open-weight decoder-only Transformer model. The instruction-tuned variant, LLaMA-3.1 8B-Instruct, is fine-tuned using supervised instruction-following data and reinforcement learning from human feedback, enhancing its alignment with user intents and safety measures. 

\mypara{Qwen-2.5 7B Base (Qwen)\cite{qwen2.5}} Qwen-2.5 7B, developed by Alibaba Cloud, is also a decoder-based model trained on an extensive 18 trillion token dataset.

\mypara{Yi-1.5 6B Base (Yi)\cite{Yi1.5}} Yi-1.5 6B, developed by 01.AI, is another decoder-based model trained on a 3.1 trillion token corpus encompassing English and Chinese texts.

\subsubsection{Datasets}
\label{sec:Data}
We evaluate unmemorization on three different types of datasets:

\mypara{Synthetic Dataset} We create a synthetic dataset consisting of 100 articles, each approximately 380 words (around 2,000 characters) in length, covering diverse topics. These articles are specifically generated to test the effectiveness of unmemorization on deeply memorized content, where the probability of reconstructing the sequence exceeds 99\%. Using synthetic data ensures that the content does not exist in the models' pretraining data, thereby providing a controlled evaluation environment.

\mypara{Organic Dataset} We evaluate on excerpts from widely-read literary works, including ``Moby Dick", ``Frankenstein", ``Adventures of Huckleberry Finn", ``Alice in Wonderland", ``Sherlock Holmes", ``A Tale of Two Cities", ``Treasure Island", ``Les Misérables", and ``Great Expectations". These works are particularly relevant as their content has been identified in publicly available models.

\mypara{Unmemorization Benchmarks} Furthermore, we evaluate \system{} on two different unmemorization/unlearning benchmarks, namely, the Copyright Takedown Evaluation (CoTaEval) \cite{wei2024CoTaEval} and the Machine Unlearning Six-Way Evaluation (MUSE) \cite{shi2024muse}. CoTaEval introduces a news dataset with 1,000 articles to forget and another 1,000 articles for the retain set. However, we do not use the retain set, as \system{} does not require any retain set. They also provide an already memorized model, which we directly use for unmemorization. This model is based on the LLaMA2 7B chat version. We use this specific version to directly compare our performance against the other techniques they report.

MUSE \cite{shi2024muse} introduces another benchmark for comparing unlearning techniques. While \system{} mainly targets verbatim memorization, we evaluate using MUSE to demonstrate \system{}'s generalizability. We focus on the News dataset of MUSE, which consists of 889 news articles in the forget set, averaging 1000 tokens each but with samples up to 3600 tokens in length—this is the longest setting we test on. Similar to CoTaEval, MUSE also presents the memorized model (LLaMA-2 7B), which we directly use for \system{}.

\mypara{Utility Benchmarks}
To assess model utility preservation, we evaluate performance on four standard benchmarks, namely HellaSwag \cite{ZHBFC19}, MMLU \cite{HBBZMSS21,HBBCLSS21}, TruthfulQA \cite{LHE22}, and Winogrande \cite{SLBC19}.

\subsection{Metrics}
We evaluate \system{} using three distinct metrics:

\mypara{Longest Common Subsequence (LCS)}
We begin by measuring the longest common subsequence of words between the generated data and the ground truth data, which the model is assumed to memorize. A higher LCS indicates a greater degree of memorization by the model.

\mypara{Edit Distance (ED)}
Next, we assess the edit distance using the Levenshtein distance, calculated based on words. A lower ED suggests a higher level of memorization by the model.
For both LCS and ED, we ignore differences in capitalization.

\mypara{ROUGE-2 Score (ROUGE-2)}
Lastly, we compute the ROUGE-2 score, which evaluates the overlap between bigrams in the generated data and the ground truth data. This metric provides another perspective on the model's ability to replicate the training data.

\begin{figure*}[!t]
\centering
\begin{subfigure}{0.49\textwidth}
\centering
\includegraphics[width=1\columnwidth]{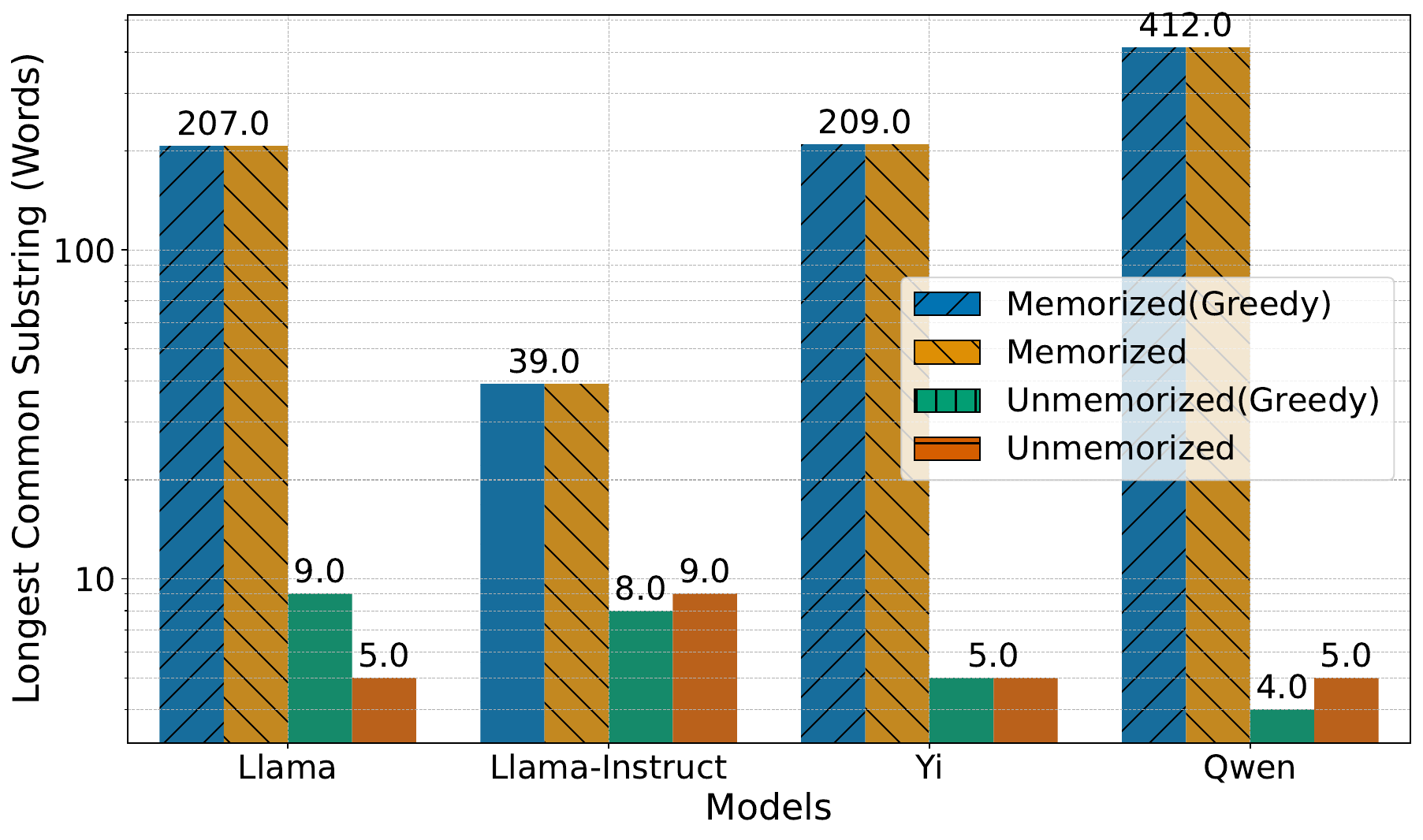}
\caption{Synthetic (LCS $\downarrow$)}
\label{fig:allModels_synthetic}
\end{subfigure}
\begin{subfigure}{0.49\textwidth}
\centering
\includegraphics[width=1\columnwidth]{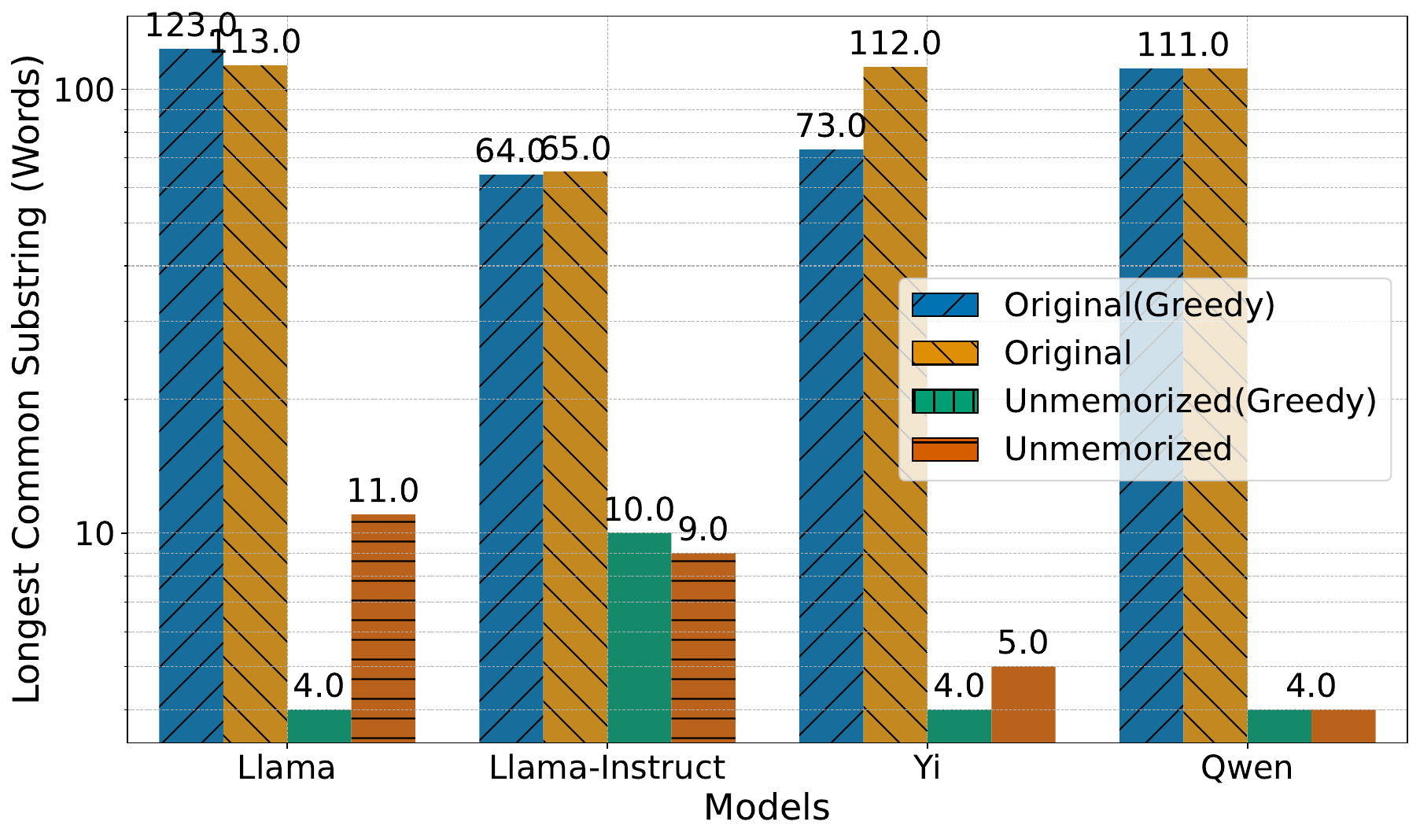}
\caption{Organic (LCS $\downarrow$)}
\label{fig:allModels_pretrain}
\end{subfigure}
\begin{subfigure}{0.49\textwidth}
\centering
\includegraphics[width=1\columnwidth]{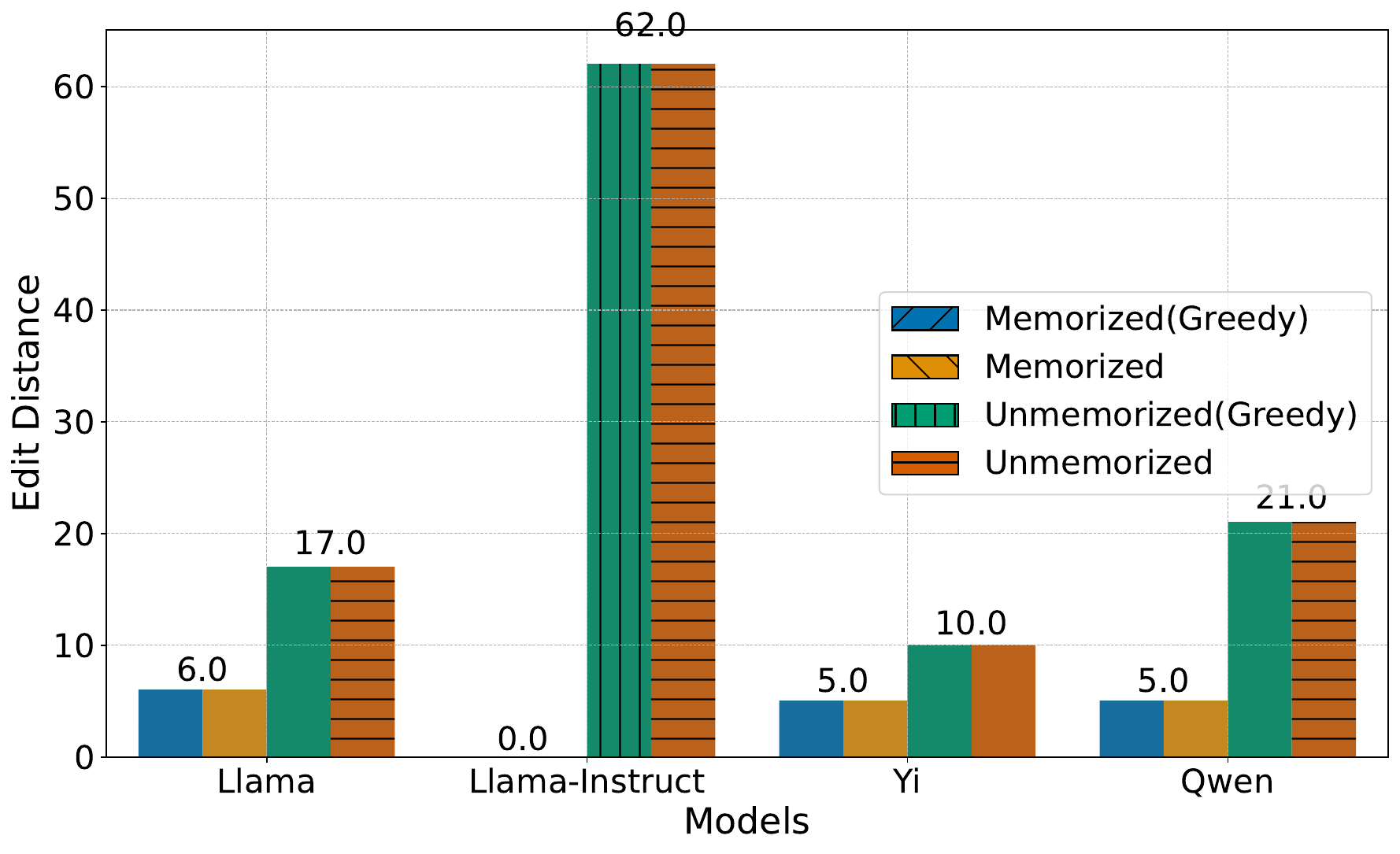}
\caption{Synthetic (ED $\uparrow$)}
\label{fig:allModels_synthetic_ed}
\end{subfigure}
\begin{subfigure}{0.49\textwidth}
\centering
\includegraphics[width=1\columnwidth]{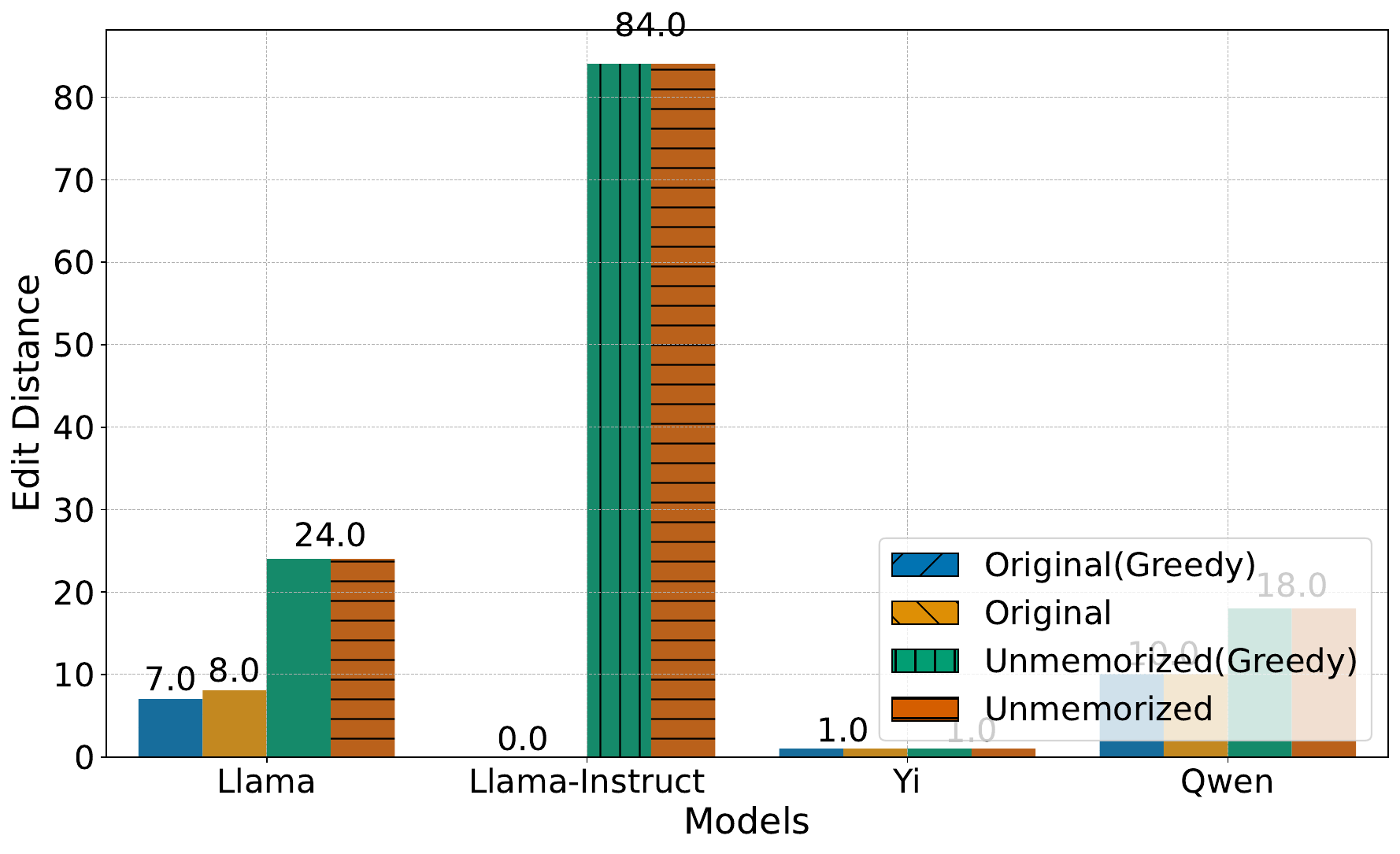}
\caption{Organic (ED $\uparrow$)}
\label{fig:allModels_pretrain_ed}
\end{subfigure}

\begin{subfigure}{0.49\textwidth}
\centering
\includegraphics[width=1\columnwidth]{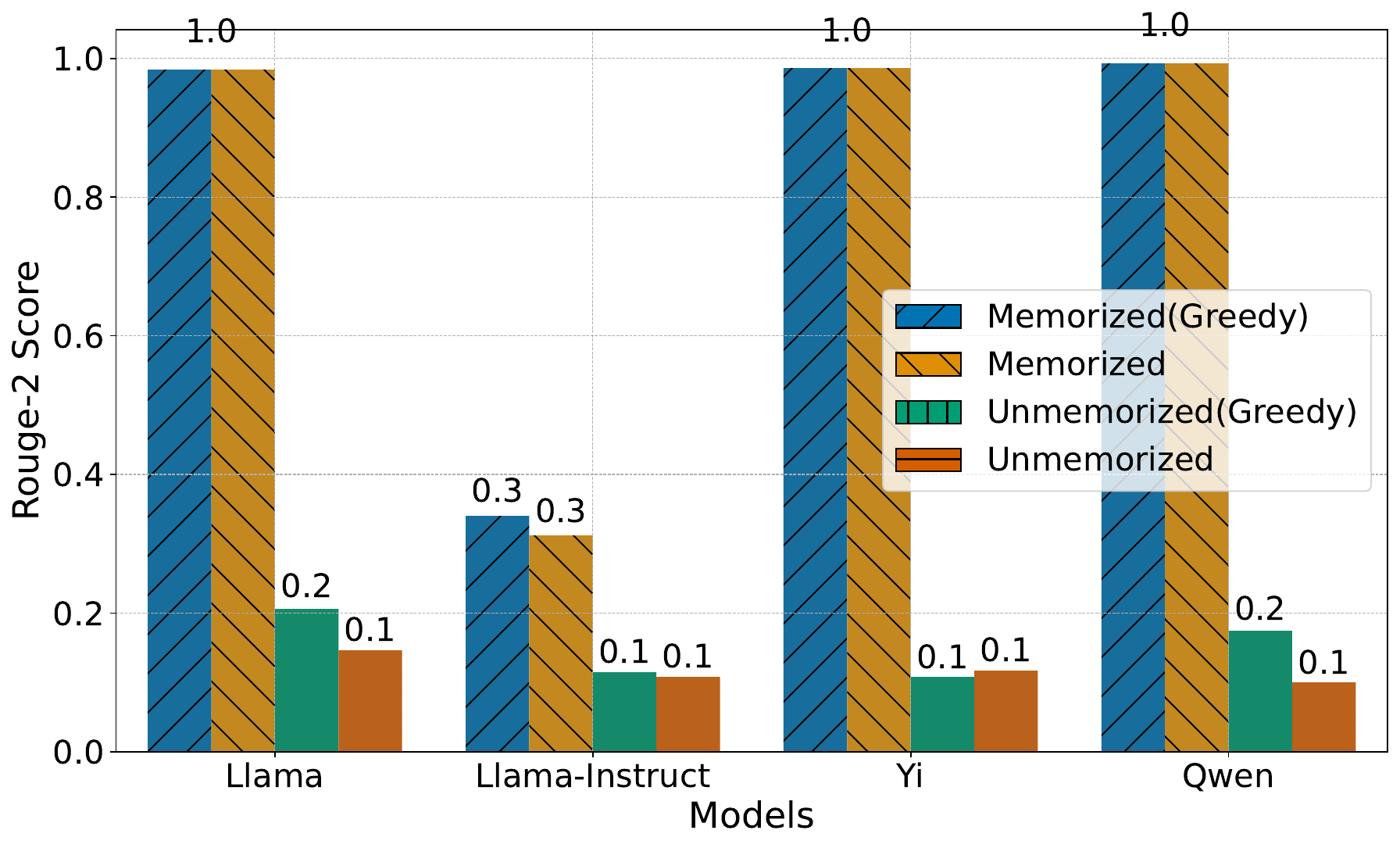}
\caption{Organic (ROUGE-2 $\downarrow$)}
\label{fig:allModels_synthetic_rouge}
\end{subfigure}
\begin{subfigure}{0.49\textwidth}
\centering
\includegraphics[width=1\columnwidth]{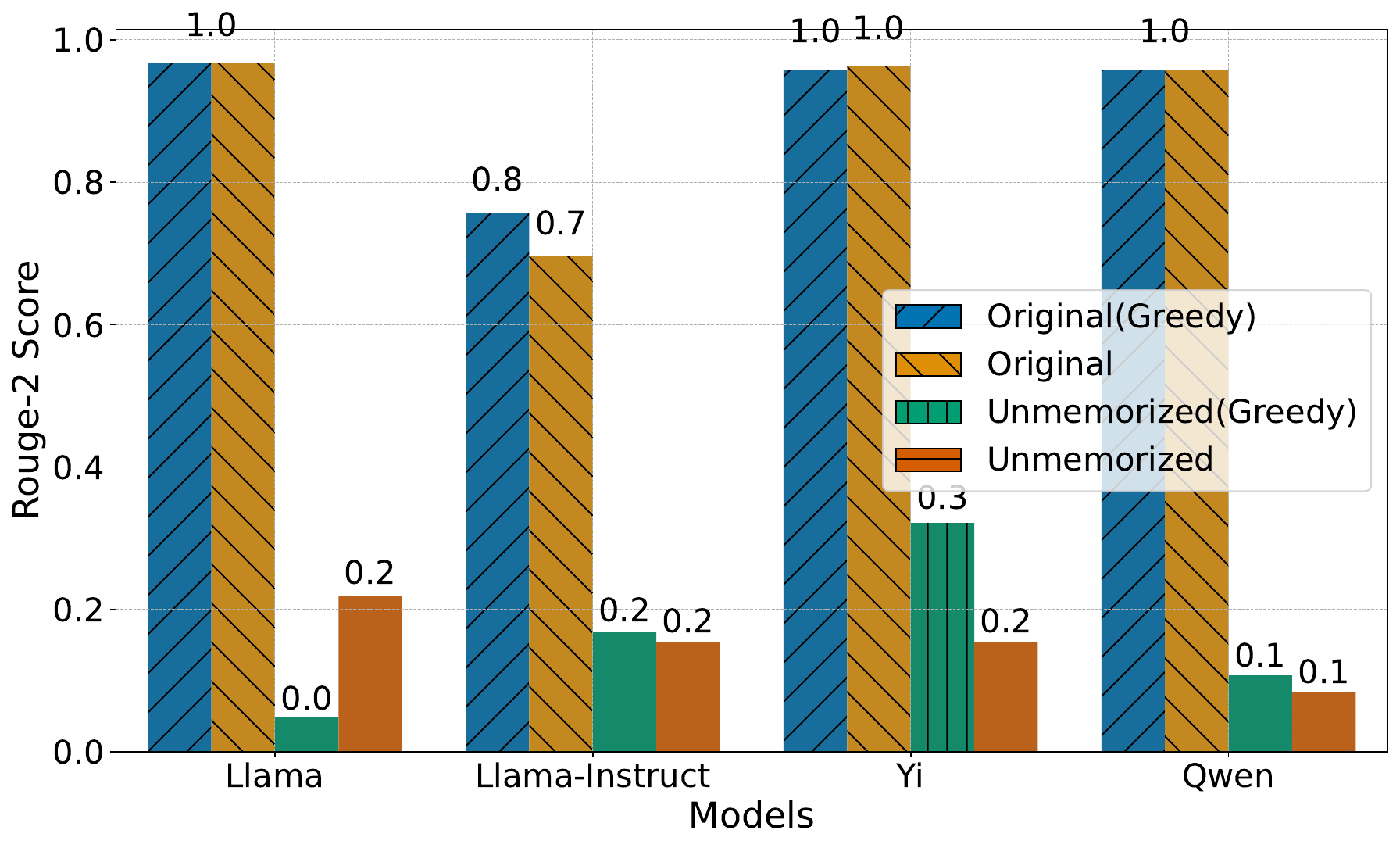}
\caption{Organic (ROUGE-2 $\downarrow$)}
\label{fig:allModels_pretrain_rouge}
\end{subfigure}
\caption{The worst-case performance of \system{} across different models, measured using Longest Common Subsequence (LCS) at the word level, Edit Distance (ED) at the word level, and ROUGE-2 score. \autoref{fig:allModels_synthetic}, \autoref{fig:allModels_synthetic_ed}, and \autoref{fig:allModels_synthetic_rouge} display the results for synthetic data, while \autoref{fig:allModels_pretrain}, \autoref{fig:allModels_pretrain_ed}, and \autoref{fig:allModels_pretrain_rouge} present the results for organic data. In each case, we consider the maximum values for LCS and ROUGE-2, and the minimum values for ED to provide a comprehensive evaluation.
}
\label{fig:allModelsRes}
\end{figure*}
\begin{figure*}[h!]  
\centering  
\begin{minipage}{0.9\textwidth}  
\textbf{Original Paragraph:}\\  
  the world. It is a way I have of driving off the spleen and regulating the circulation. Whenever I find myself growing grim about the mouth; whenever it is a damp, drizzly November in my soul; whenever I find myself involuntarily pausing before coffin warehouses, and bringing up the rear of every funeral I meet; and especially whenever my hypos get such an upper hand of me, that it requires a strong moral principle to prevent me from deliberately stepping into the street, and methodically knocking people’s
\end{minipage}  

\begin{minipage}{0.9\textwidth}  
\textbf{Generated Paragraph After Unmemoriazton:}\\  
  the\textcolor{red}{y can tell you, too, that the life is one of hard work, long hours, and a very uncertain future. The life of a sailor is not for a man who is averse to work. But for those who have the sea in their blood, and [$\cdots$]
} 
\end{minipage}  
\caption{A demonstration of how the model diverges from copyrighted content following unmemorization by the system. The differences are highlighted in red in the generated text.}  
  \label{fig:compText}
  \end{figure*}

\begin{table*}[h!]  
\centering  
\caption{Benchmarking \system{} when Targeting Synthetic and Organic Data.}  
\resizebox{\linewidth}{!}{
\begin{tabular}{llcccccccc}    
\toprule  
\textbf{Model} & \textbf{Type} & \multicolumn{2}{c}{\textbf{MMLU}} & \multicolumn{2}{c}{\textbf{TruthfulQA}} & \multicolumn{2}{c}{\textbf{HellaSwag}} & \multicolumn{2}{c}{\textbf{Winogrande}} \\  
 & & \textbf{Organic} & \textbf{Synthetic} & \textbf{Organic} & \textbf{Synthetic} & \textbf{Organic} & \textbf{Synthetic} & \textbf{Organic} & \textbf{Synthetic} \\  
\midrule  

\multirow{3}{*}{\textbf{Llama }} & Mem & 0.6324 & 0.6322 & 0.4519 & 0.5110 & 0.6007 & 0.6109 & 0.7388 & 0.7174 \\
& UnMem & 0.6337 & 0.6281 & 0.4551 & 0.5044 & 0.6010 & 0.6123 & 0.7419 & 0.7245 \\
& Difference & $\uparrow$ 0.21\% & $\downarrow$ 0.65\% & $\uparrow$ 0.71\% & $\downarrow$ 1.29\% & $\uparrow$ 0.05\% & $\uparrow$ 0.23\% & $\uparrow$ 0.42\% & $\uparrow$ 0.99\% \\
 \midrule 
\multirow{3}{*}{\textbf{Llama-Inst }} & Mem & 0.6790 & 0.6812 & 0.5402 & 0.5387 & 0.5905 & 0.5937 & 0.7380 & 0.7340 \\
 & UnMem & 0.6773 & 0.6787 & 0.5425 & 0.5348 & 0.5939 & 0.5970 & 0.7356 & 0.7403 \\
& Difference & $\downarrow$ 0.25\% & $\downarrow$ 0.37\% & $\uparrow$ 0.43\% & $\downarrow$ 0.72\% & $\uparrow$ 0.58\% & $\uparrow$ 0.56\% & $\downarrow$ 0.33\% & $\uparrow$ 0.86\% \\
  \midrule 
\multirow{3}{*}{\textbf{Yi}} & Mem & 0.6242 & 0.6226 & 0.4409 & 0.4529 & 0.5671 & 0.5800 & 0.7206 & 0.7056 \\
 & UnMem & 0.6227 & 0.6202 & 0.4409 & 0.4597 & 0.5657 & 0.5819 & 0.7174 & 0.7174 \\
& Difference & $\downarrow$ 0.24\% & $\downarrow$ 0.39\% & $\downarrow$ 0.00\% & $\uparrow$ 1.50\% & $\downarrow$ 0.25\% & $\uparrow$ 0.33\% & $\downarrow$ 0.44\% & $\uparrow$ 1.67\% \\
 \midrule 
\multirow{3}{*}{\textbf{Qwen}} & Mem & 0.7190 & 0.7192 & 0.5638 & 0.5584 & 0.6006 & 0.6037 & 0.7340 & 0.7143 \\
 & UnMem & 0.7161 & 0.7191 & 0.5690 & 0.5715 & 0.5975 & 0.6176 & 0.7230 & 0.6930 \\
& Difference & $\downarrow$ 0.40\% & $\downarrow$ 0.01\% & $\uparrow$ 0.92\% & $\uparrow$ 2.35\% & $\downarrow$ 0.52\% & $\uparrow$ 2.30\% & $\downarrow$ 1.50\% & $\downarrow$ 2.98\% \\
\bottomrule  
\end{tabular}  
}  
\label{tab:benchmarkAllSyn}
\end{table*}  
\begin{figure*}[!t]
\centering
\begin{subfigure}{0.49\textwidth}
\centering
\includegraphics[width=1\columnwidth]{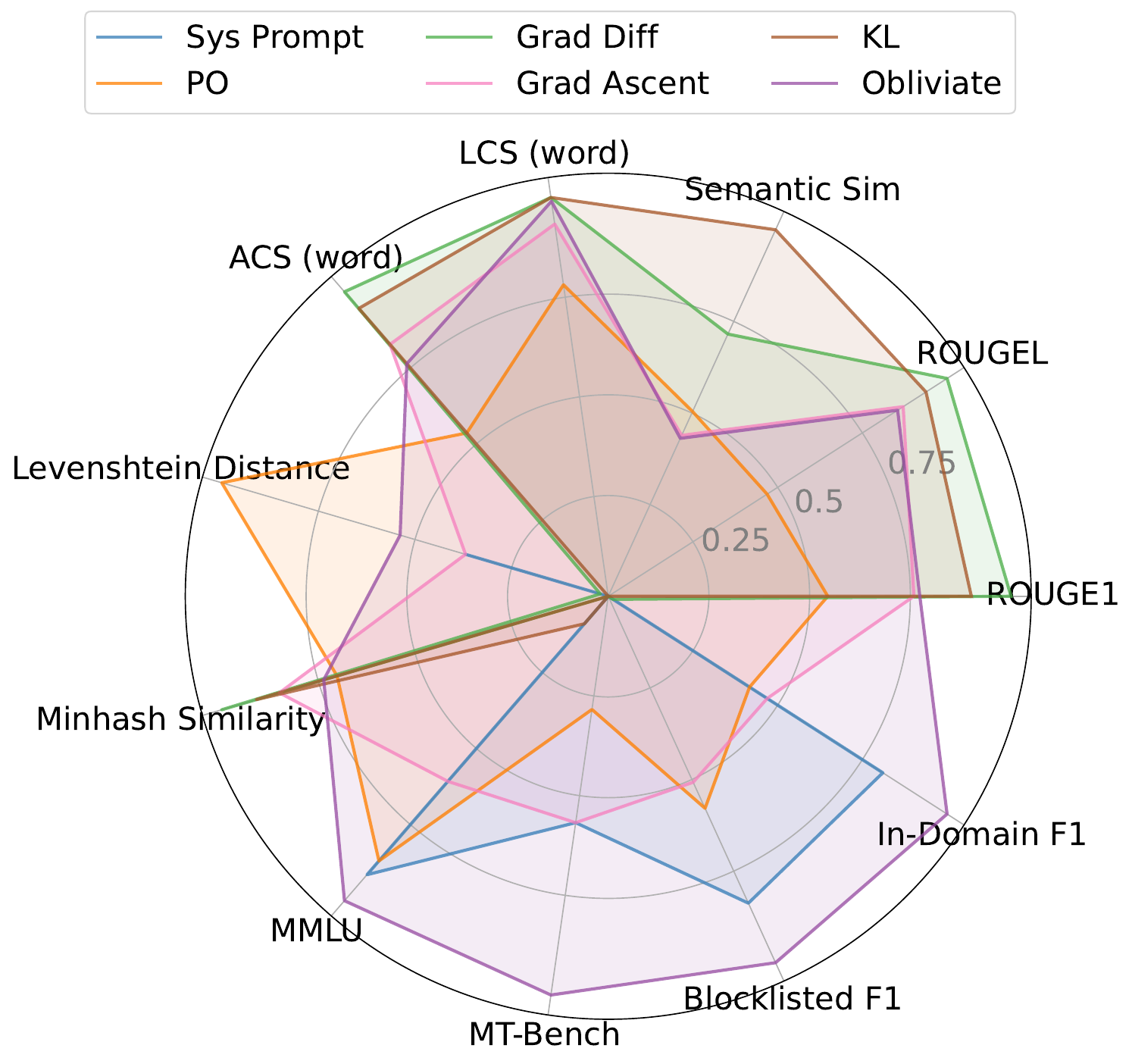}
\caption{
Max.}
\label{fig:CoTaEval_max}
\end{subfigure}
\begin{subfigure}{0.49\textwidth}
\centering
\includegraphics[width=1\columnwidth]{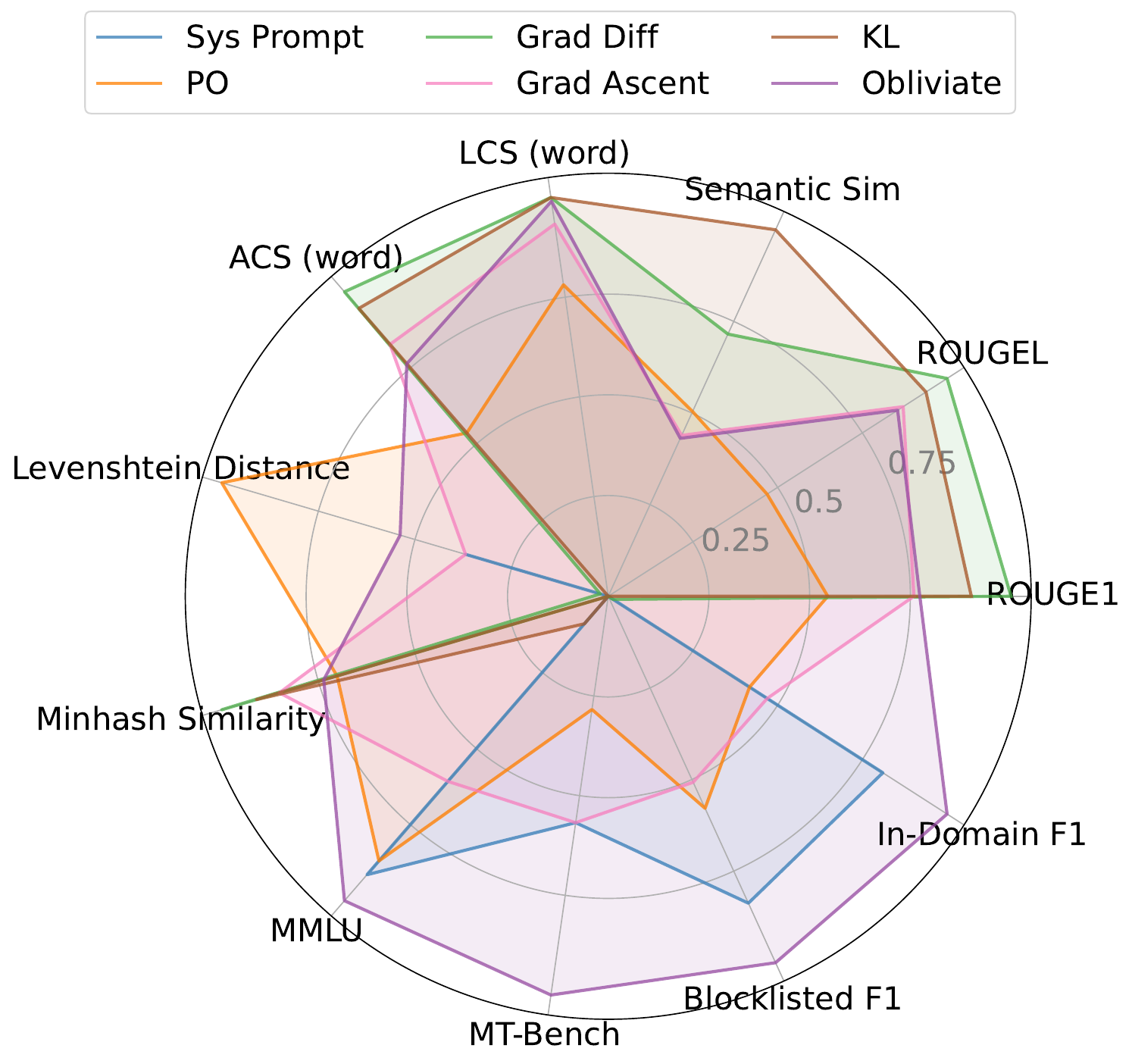}
\caption{
Mean.}
\label{fig:CoTaEval_mean}
\end{subfigure}
\caption{Comparison of different copyright takedown methods using CoTaEval benchmarks for both mean and maximum values. Values are normalized, and for ROUGE-1, ROUGE-L, semantic similarity, LCS (word), ACS (word), and Minhash similarity, the values are inverted to ensure that higher values (closer to the boundary) indicate better performance.}
\label{fig::CoTaEval}
\end{figure*}
\begin{figure}[!t]
\centering
\includegraphics[width=1\columnwidth]{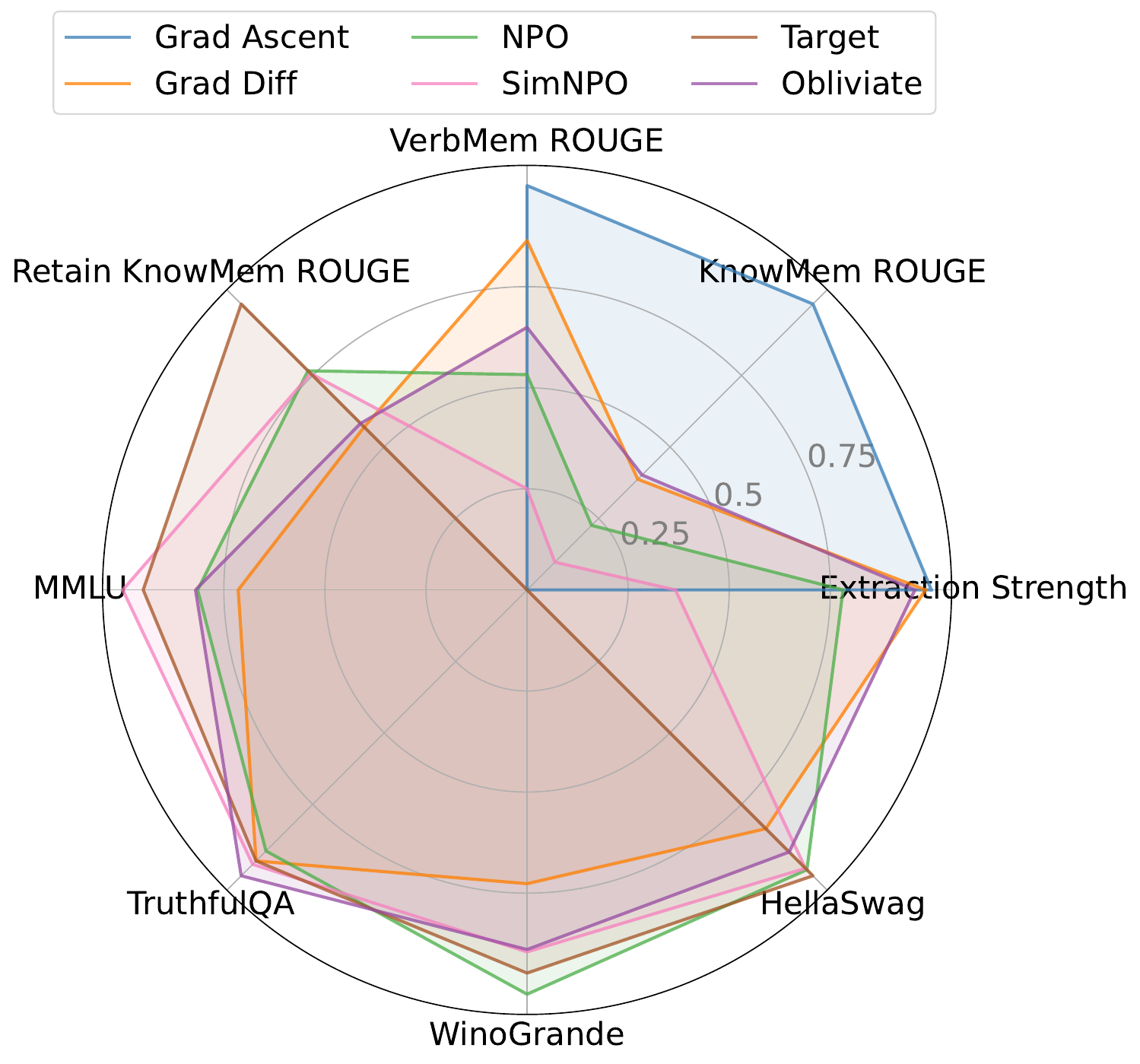}
\caption{
Evaluation result on the MUSE benchmark.
}
\label{fig:muse}
\end{figure}

\begin{figure*}[!t]
\centering
\begin{subfigure}{0.49\textwidth}
\centering
\includegraphics[width=1\columnwidth]{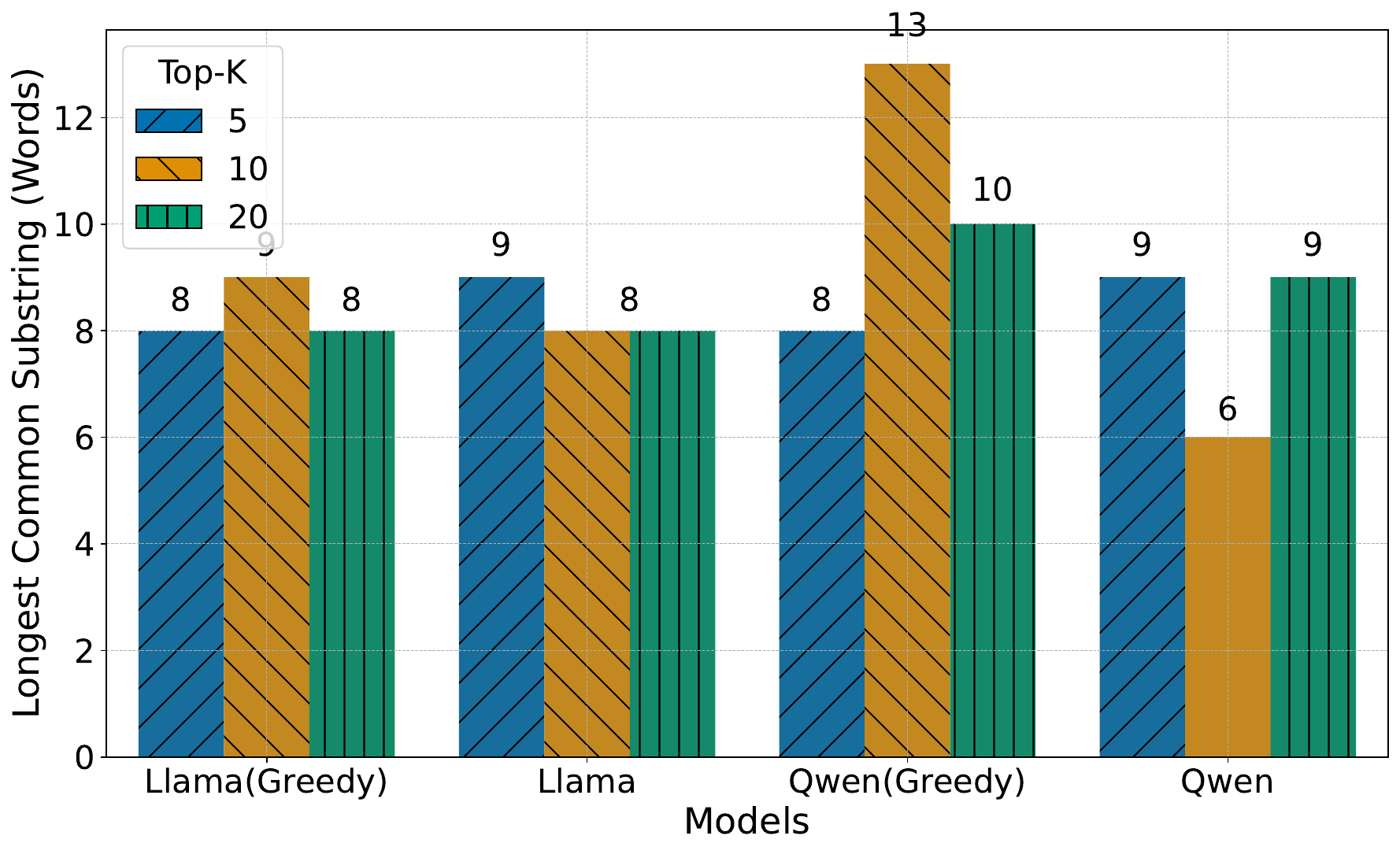}
\caption{
Synthetic (LCS $\downarrow$).}
\label{fig:topK_synthetic}
\end{subfigure}
\begin{subfigure}{0.49\textwidth}
\centering
\includegraphics[width=1\columnwidth]{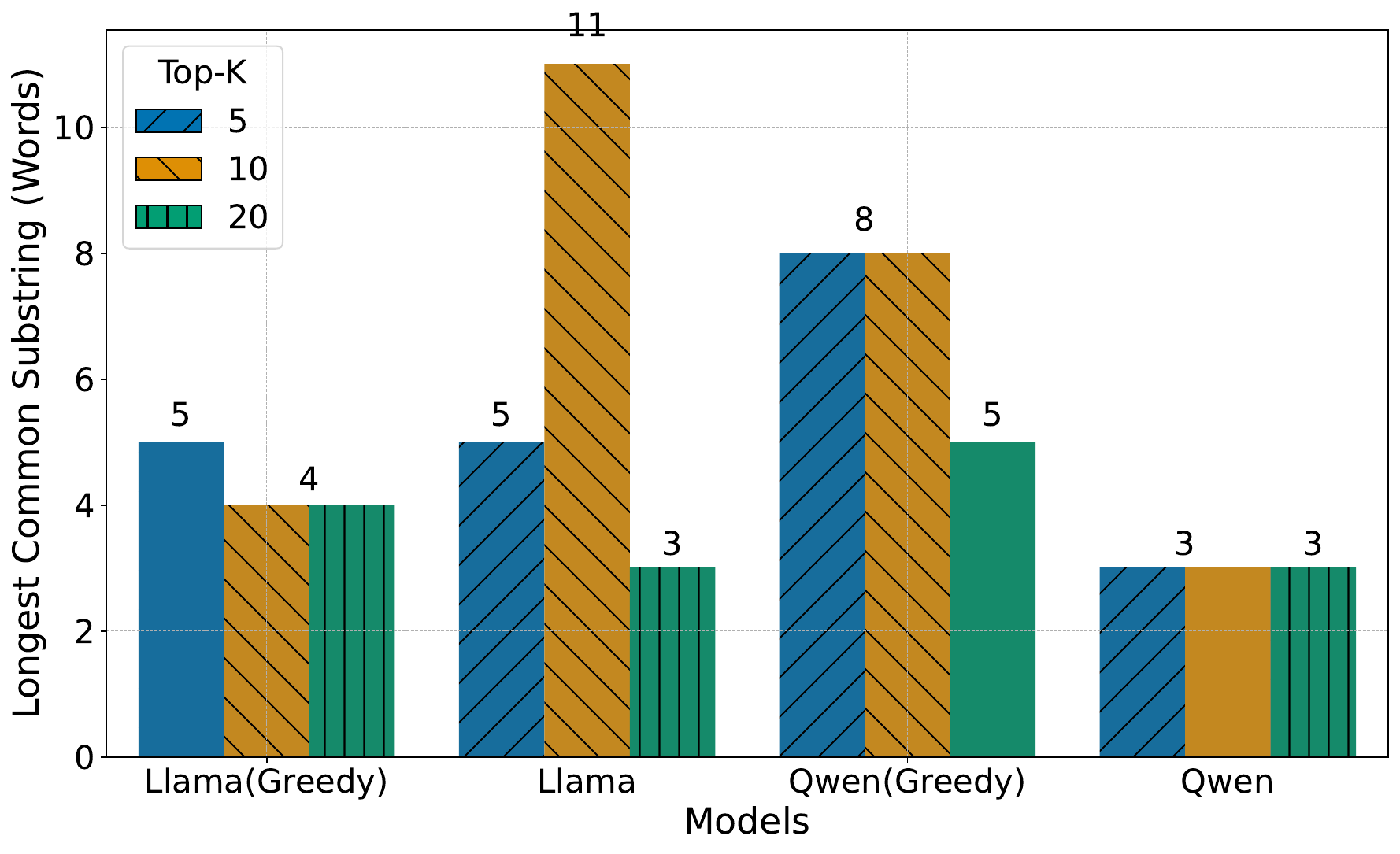}
\caption{
Organic (LCS $\downarrow$).}
\label{fig:topK_org}
\end{subfigure}
\caption{
The effect of varying the Top-k probabilities for $\floss$ and $\mloss$ using organic and synthetic data.}
\label{fig:topK}
\end{figure*}

\subsection{Results}
We first evaluate \system{} through two main experiments: (1) unmemorizing synthetically memorized data (simulating deeply memorized), and (2) unmemorizing organically memorized data present in the pre-trained models. For each scenario, we assess both the effectiveness of \system{} memorization and its effect on model utility.

\mypara{Unmemorizing Synthetic Data}
We first finetune the target models on 100 samples using our synthetic dataset until deep memorization is achieved, i.e., tokens are memorized up to at least 99\% confidence. Intuitively, we finetune the model till its ability to reproduce the complete article given a prefix. We refer to these as ``Memorized'' models.

Next, we apply \system{} with a token stride of 4, modifying the token following every set of four tokens, and a modification rate of one token per modification. For the KL-divergence, we consider the top-10 probabilities. We implement early stopping  when the forget loss  falls below a predetermined threshold, which correlates with target tokens achieving low probabilities.

To comprehensively evaluate memorization, we consider different testing settings that vary prefix offset positions, i.e., where the prefix starts, ranging from 0 to 200 tokens at 50-token intervals, and prefix lengths ranging from 8 to 20 tokens. We examine both greedy decoding (Greedy) and temperature-based sampling (temperature = 0.6, the default for LLaMA). We consider the \textbf{worst-case scenario} in evaluation by considering the absolute maximum for all sentences and inputs for the Longest Common Subsequence (LCS) and ROUGE-2, while considering the minimum Edit Distance (ED). Intuitively, this means that if \textbf{ALL} inputs were 100\% forgotten except for one, this evaluation method would completely indicate it as a failure. For completeness, we also show the mean results; however, we believe the maximum/minimum metrics are the best indicators of how successful an unmemorization technique is. Finally, for the instruct version of the model, we precede the prefix with the following instruction: ``Generate the entire rest of this text from the start, continuing until you reach the end: ''.

\autoref{fig:allModelsRes} demonstrates the efficacy of our approach across multiple evaluation metrics and model architectures. The Longest Common Subsequence (LCS) analysis reveals a dramatic reduction in memorization: while baseline models reproduce substantial portions of training data—with memorized sequences, for instance, Qwen (\autoref{fig:allModels_synthetic}) generates LCS of size 412 words, which is reduced to 4-5 words after applying \system{}, achieving over two orders of magnitude improvement. Similar patterns are observed across different models and datasets. This is also demonstrated by the Edit Distance results, which significantly increase after applying \system{} for all models and datasets, and by the ROUGE-2 scores, which provide additional confirmation, demonstrating the significant unmemorization achieved by \system{}.

\mypara{Unmemorizing Organic Data}
Our second experiment focuses on organically memorized data present in publicly released models, i.e., without any additional fine-tuning from us. Using the dataset described in \autoref{sec:Data}, we apply the same \system{} parameters and evaluation methodology as in the synthetic experiments. Results demonstrate comparable effectiveness in removing organic memorization, with LCS (Figure \ref{fig:allModels_pretrain}) reduced to 2-5 words. Both ED (Figure \ref{fig:allModels_pretrain_ed}) and ROUGE-2 (Figure \ref{fig:allModels_pretrain_rouge}) metrics show reduction of memorization to negligible levels. Additionally, we present an example in \autoref{fig:compText} illustrating how text generation diverges after unmemorization, resulting in completely different content.\mypara{\system{} Effect on Utility}
We assess the impact on model utility by comparing \system{}-unmemorized models against their respective original models across standard benchmarks. \autoref{tab:benchmarkAllSyn} demonstrates the stability of \system{}, with performance degradation consistently remaining minimal across all evaluated tasks and model architectures.

For example, for the Llama model family, MMLU scores show an improvement of 0.21$\uparrow$\% and a degradation of only 0.63$\downarrow$\% for organic and synthetic data, respectively. TruthfulQA exhibits a similar improvement for organic data at 0.71$\uparrow$\%, and slightly larger but still modest degradations at 1.29$\downarrow$\%. HellaSwag performance remains virtually unchanged with a change of only 0.05$\uparrow$\% and 0.23$\downarrow$\%, and Winogrande shows an improvement of 0.42$\uparrow$\% for organic data and a degradation of 0.99$\downarrow$\% for synthetic data. Similarly, Llama-Inst maintains strong performance with MMLU degradations of 0.25$\downarrow$\% and 0.37$\downarrow$\%, TruthfulQA changes of 0.43$\uparrow$\% and 0.72$\downarrow$\%, HellaSwag an improvement of 0.58$\uparrow$\% and 0.56$\uparrow$\%, and Winogrande a degradation of 0.33$\downarrow$\% and an improvement of 0.86$\uparrow$\%.

In general, across all 32 evaluation scenarios (4 models × 4 benchmarks × 2 datasets), the maximum degradation observed is 2.98\% on Winogrande for Qwen with synthetic data, while the majority of cases show degradations well below 1\% and some an improvement. This consistent performance across diverse model architectures—from base models (Llama, Yi, Qwen) to instruction-tuned variants (Llama-Inst)—demonstrates that \system{} successfully preserves model utility while effectively removing memorized content.

These results establish that \system{} achieves strong unmemorization performance without sacrificing model capabilities, presenting an efficient approach for selective knowledge removal in large language models. The consistent utility preservation across various model sizes, architectures, and both synthetic and organic memorization scenarios show the robustness and practical applicability of our approach across different deployment contexts.
\begin{table*}[ht]
    \centering
    \begin{tabular}{@{}lrrrrrrrrrrrr@{}}
        \toprule
        \textbf{Method} & \textbf{ROUGE1} & \textbf{ROUGEL} & \textbf{Sem Sim} & \textbf{LCS} & \textbf{ACS} & \textbf{LD} & \textbf{Minhash} & \textbf{MMLU} & \textbf{MT‑Bench} & \textbf{Blocked F1} & \textbf{In‑Domain F1} \\
        \midrule
        Sys Prompt   & 0.839 & 0.801 & 0.930 & 119  & 124  & 1107 & 0.812 & 0.340 & 0.330  & 0.334 & 0.330 \\
        PO           & 0.579 & 0.515 & 0.868 &  37  &  65  & 1230 & 0.445 & 0.331 & 0.240  & 0.283 & 0.247 \\
        Grad Diff    & 0.362 & 0.191 & 0.842 &  14  &  14  & 1039 & 0.289 & 0.158 & 0.150  & 0.169 & 0.162 \\
        Grad Ascent  & 0.477 & 0.270 & 0.876 &  21  &  33  & 1107 & 0.367 & 0.279 & 0.330  & 0.269 & 0.258 \\
        KL           & 0.409 & 0.229 & 0.807 &  14  &  20  & 1035 & 0.336 & 0.176 & 0.150  & 0.169 & 0.159 \\
        Obliviate    & 0.470 & 0.280 & 0.877 &  15  &  40  & 1140 & 0.427 & 0.357 & 0.467  & 0.366 & 0.370 \\
        \bottomrule
    \end{tabular}
    \caption{CoTaEval results.}
    \label{table:cotaevalMax}
\end{table*}

\subsection{Generalizability and Comparison}

Next, we demonstrate the generalizability of \system{} by evaluating it against two benchmarks: CoTaEval~\cite{wei2024CoTaEval} and MUSE~\cite{shi2024muse}.

\mypara{CoTaEval}
We begin with CoTaEval, which encompasses multiple copyright‐enforcement mechanisms, including both training‐time approaches (similar to \system{}) and test‐time approaches. In this work, we focus primarily on the comparison with training‐time techniques that are most similar to \system{}, namely Gradient Ascent (Grad Ascent)\cite{Thudi22gradAscent}, Gradient Difference (Grad Diff)\cite{liu202graddiff}, KL Minimization (KL)\cite{golatkar2020KL}, and Preference Optimization (PO)\cite{rafailov2024idk}. We also include a simple unmemorization baseline that explicitly states the forget request in the system prompt (Sys Prompt). All methods are applied to a Llama‐2 7B chat model that has been memorized on the NewsQA dataset (it is provided by CoTaEval). The NewsQA forget set consists of 1,000 inputs to forget, which further tests the robustness of each method when scaling to larger datasets.

CoTaEval evaluates forgetting‐effectiveness using the following metrics:
\begin{itemize}
\item \textbf{ROUGE-1 / ROUGE-L}\cite{Lin2004rouge}: unigram and longest‐common‐subsequence overlap between the generated output and the copyrighted/target content to forget;
\item \textbf{LCS (character) / LCS (word)}: longest common subsequence measured at the character and word levels;
\item \textbf{ACS (word)}: accumulated common subsequences in words;
\item \textbf{Levenshtein Distance}\cite{levenshtein1966binary}: the edit distance between the generated output and the blocklisted content;
\item \textbf{Semantic Similarity}: cosine similarity between the generated output and the copyrighted/target content to forget;
\item \textbf{MinHash Similarity}~\cite{Broder1997minHash}: the Jaccard‐approximation similarity via MinHash sketches.
\end{itemize}

To assess overall model utility, CoTaEval additionally measures: \textbf{MMLU} (Multi‐Task Language Understanding) accuracy
and \textbf{MT‐Bench}~\cite{Zheng2023MTBench} scores.

Finally, to quantify in‐domain knowledge retention for inputs not in the forget set, CoTaEval reports word‐level \textbf{F1} on (i) \textbf{blocklisted} content and (ii) \textbf{in‐domain} content that should \emph{not} be forgotten.

\autoref{fig:CoTaEval_max} shows the comparison of all unmemorization techniques on CoTaEval using maximum metric values, and \autoref{fig:CoTaEval_mean} shows the same comparison using mean metric values (the utility metrics remain identical in both figures). Both figures are normalized and adjusted—so that for ROUGE-1, ROUGE-L, Semantic Similarity (Semantic Sim), LCS(word), ACS(word) and MinHash Similarity, lower raw scores are inverted (since lower values are better for these metrics) to yield higher normalized values, i.e., closer to the boundary is always best. Grad Diff achieves the best performance with respect to ROUGE-1 (0.362 vs. 0.470 for \system{}), ROUGE-L (0.191 vs. 0.280 for \system{}), LCS(word) (14 vs. 15 for \system{}), ACS(word) (14 vs. 40) and MinHash (0.289 vs. 0.427), while KL Minimization outperforms \system{} only on Semantic Similarity (raw 0.807 vs. 0.877 ) and PO outperforms \system{} only on Levenshtein Distance (raw 1,230 vs. 1,140). In contrast, \system{} dominates all utility metrics—MMLU (0.357), MT-Bench (4.67), Blocklisted F1 (0.366) and In-Domain F1 (0.370)—demonstrating the best overall utility. This result shows that \system{} strikes a more balanced trade-off by combining near–state-of-the-art forgetting performance with significantly higher utility. We present the concrete numbers for the maximum evaluation of CoTaEval in \autoref{table:cotaevalMax}.

\mypara{MUSE} Second, we evaluate MUSE on the News dataset (distinct from the dataset used in CoTaEval). This dataset comprises 889 samples to forget, averaging approximately 1000 tokens in length with samples as long as 3600 tokens. Due to memory constraints, we split the inputs into segments of at most 512 tokens, resulting in an increased number of forget‐inputs. The MUSE inputs are heavily formatted with punctuations, which had an effect on the performance of \system{}. To address this, we simply modified the stride selection to ignore punctuations, i.e., if punctuations are selected as target tokens to unmemorize. Instead, the \system{} would search for the next token, first by going to the previous tokens, and if not found, then searching the next tokens. After selecting a new target token, the stride would start from this new position and continue normally.

The MUSE benchmark also employs the ROUGE‐L metric in two distinct modes: \textbf{No Verbatim Memorization} (VerbMem ROUGE), which ensures that the model does not reproduce the exact input, thus measuring memorization; and \textbf{No Knowledge Memorization} (KnowMem ROUGE), which requires the model to erase all knowledge from the input, testing complete unlearning—a capability \system{} does not aim to provide. Additionally, MUSE uses \emph{Extraction Strength} to quantify how closely the model’s predictions align with ground‑truth labels. To assess performance on data that should not be forgotten, MUSE also computes ROUGE‐L on the ``retain'' subset (Retain KnowMem ROUGE). Beyond MUSE's utility evaluation, we evaluate all techniques across four additional benchmarks: MMLU, TruthfulQA, WinoGrande, and HellaSwag. We do not use MT-Bench here as it is costly to run since it uses GPT-4 as a judge.

MUSE considers gradient‐based methods analogous to those in CoTaEval (Grad Ascent and Grad Difference), as well as negative preference optimization (NPO)~\cite{zhang2024NPO}, SimNPO~\cite{fan2025SimNPO}, and the original memorized model (Target) as the unlearning techniques.

We follow the same normalization procedure from CoTaEval. We invert VerbMem ROUGE, KnowMem ROUGE, and Extraction Strength before plotting, since lower scores indicate better performance for these metrics. The results, shown in\autoref{fig:muse}, reveal that, similar to CoTaEval, Grad Difference most closely approaches \system{}’s overall performance; however, \system{} still achieves higher utility, nearly matching the Target model. This demonstrates that \system{} can effectively unmemorize the target inputs while maintaining strong utility. Finally, it is important to recap that unlike other methodologies, \system{} does not require a retain dataset, reducing the number of assumptions needed for execution.
 \begin{figure*}[!t]
\centering
\begin{subfigure}{0.49\textwidth}
\centering
\includegraphics[width=1\columnwidth]{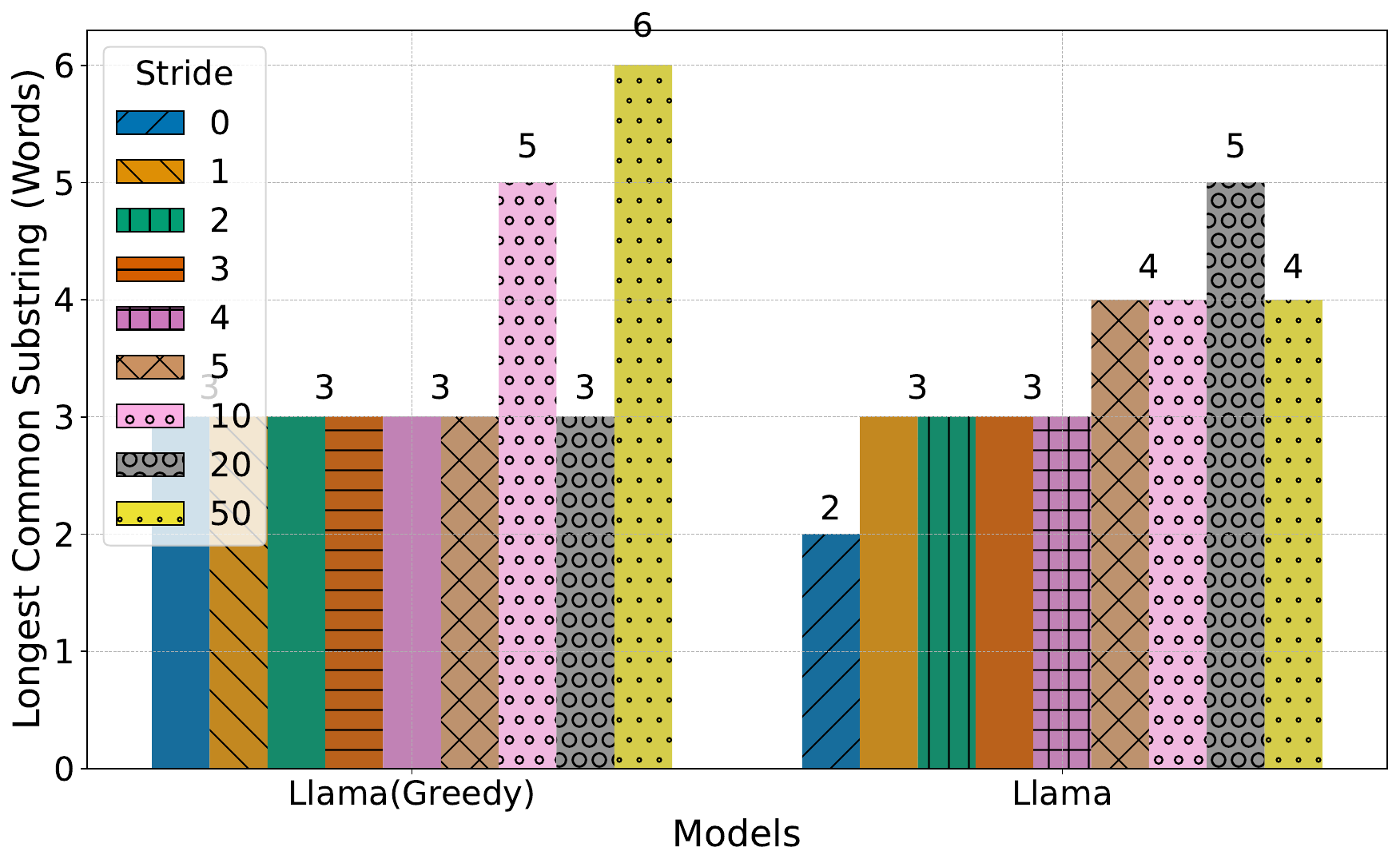}
\caption{Synthetic (LCS $\downarrow$)}
\label{fig:strideSyn}
\end{subfigure}
\begin{subfigure}{0.49\textwidth}
\centering
\includegraphics[width=1\columnwidth]{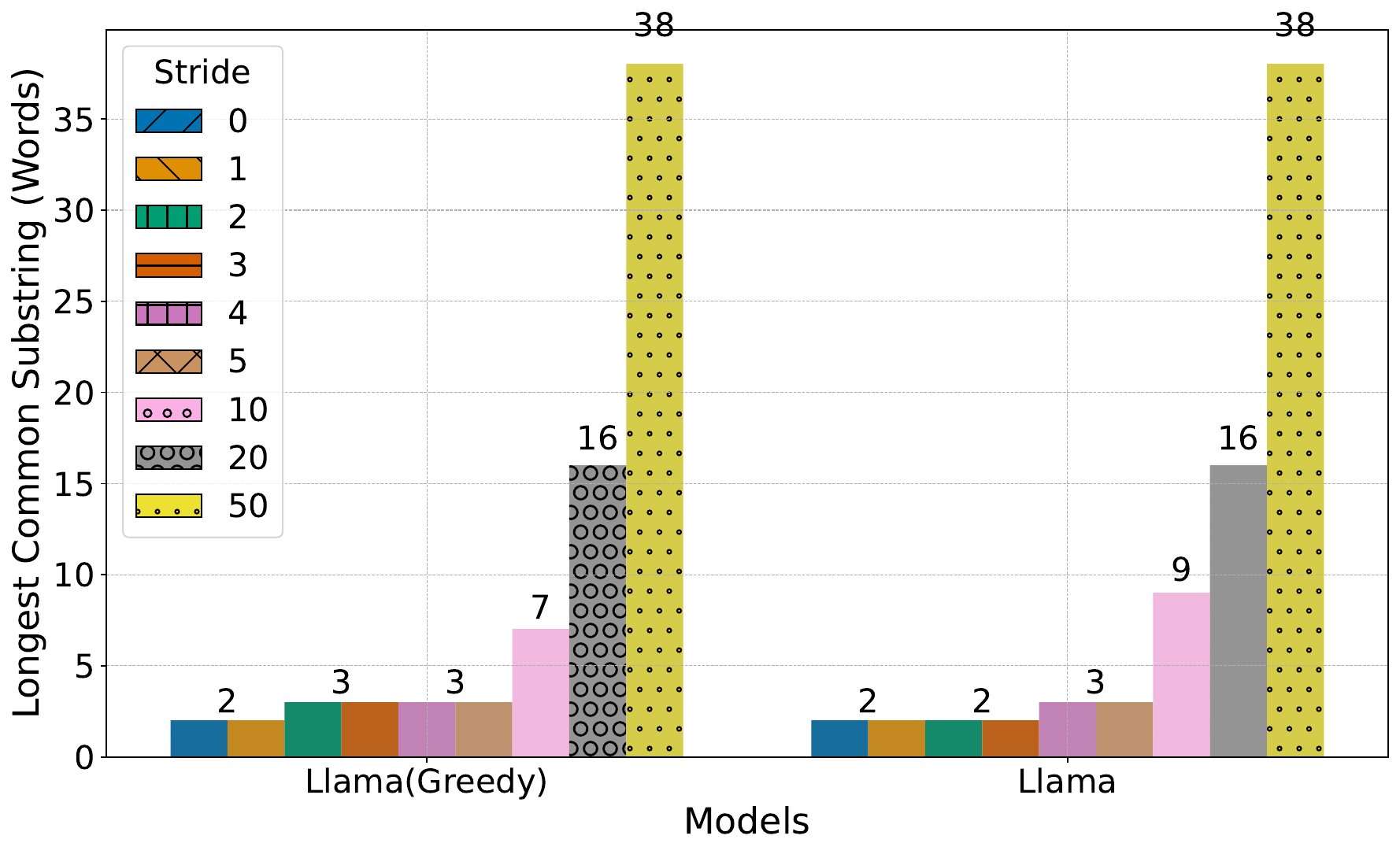}
\caption{Organic (LCS $\downarrow$)}
\label{fig:strideOrg}
\end{subfigure}
\begin{subfigure}{0.49\textwidth}
\centering
\includegraphics[width=1\columnwidth]{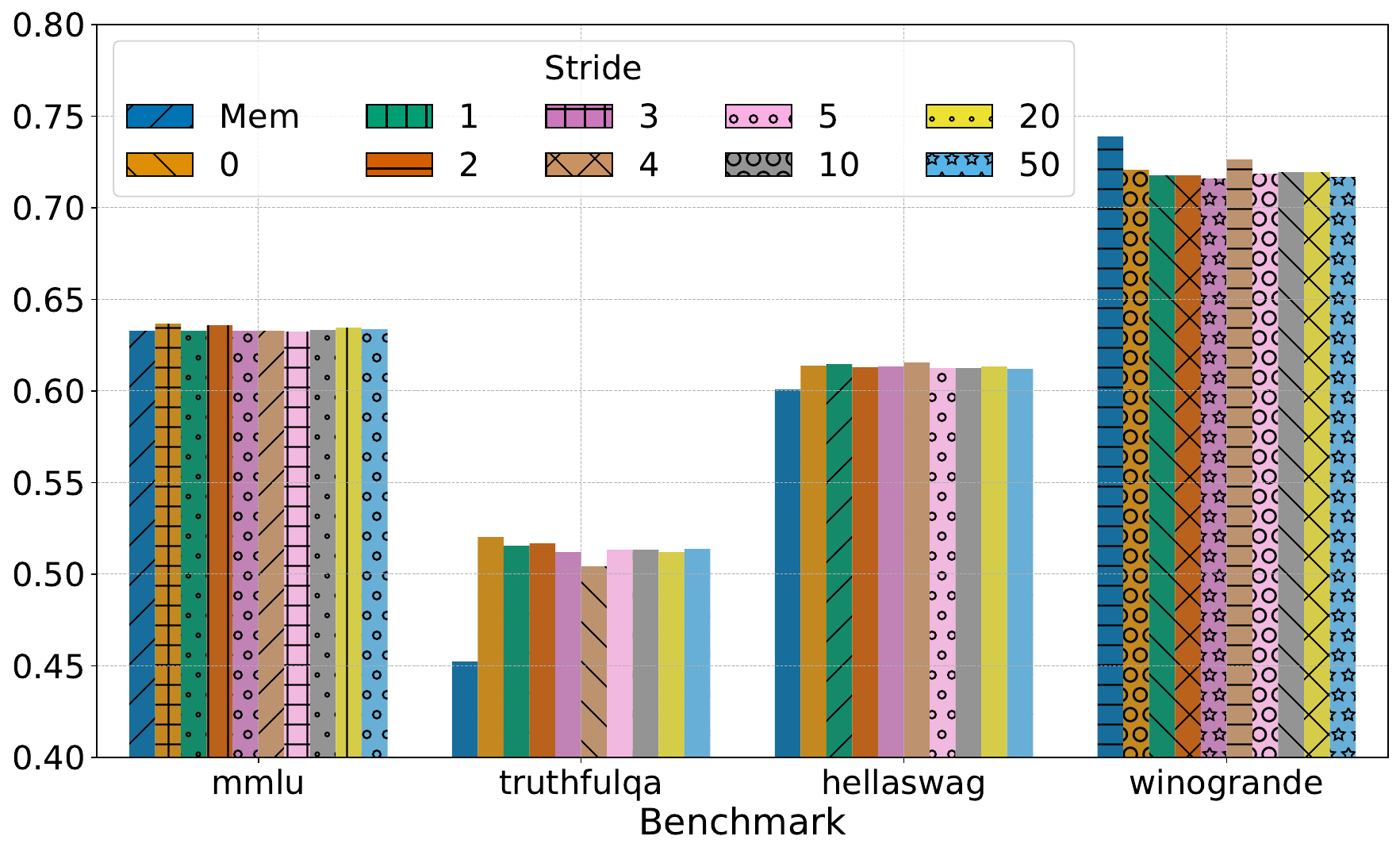}
\caption{Benchmark Llama Synthetic}
\label{fig:strideBenchLlamaSyn}
\end{subfigure}
\begin{subfigure}{0.49\textwidth}
\centering
\includegraphics[width=1\columnwidth]{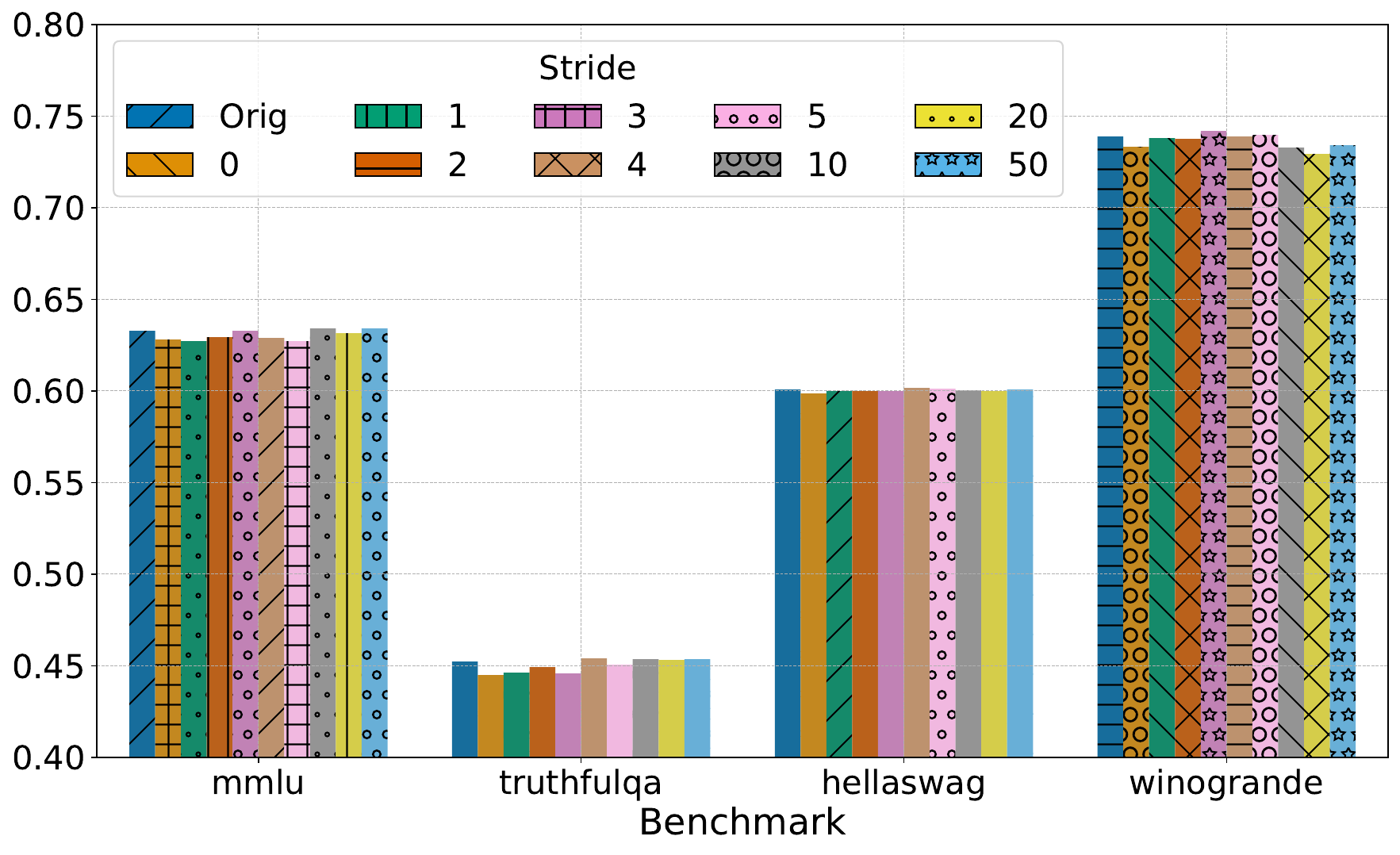}
\caption{Benchmark Llama Organic}
\label{fig:strideBenchLlamaOrg}
\end{subfigure}
\begin{subfigure}{0.49\textwidth}
\centering
\includegraphics[width=1\columnwidth]{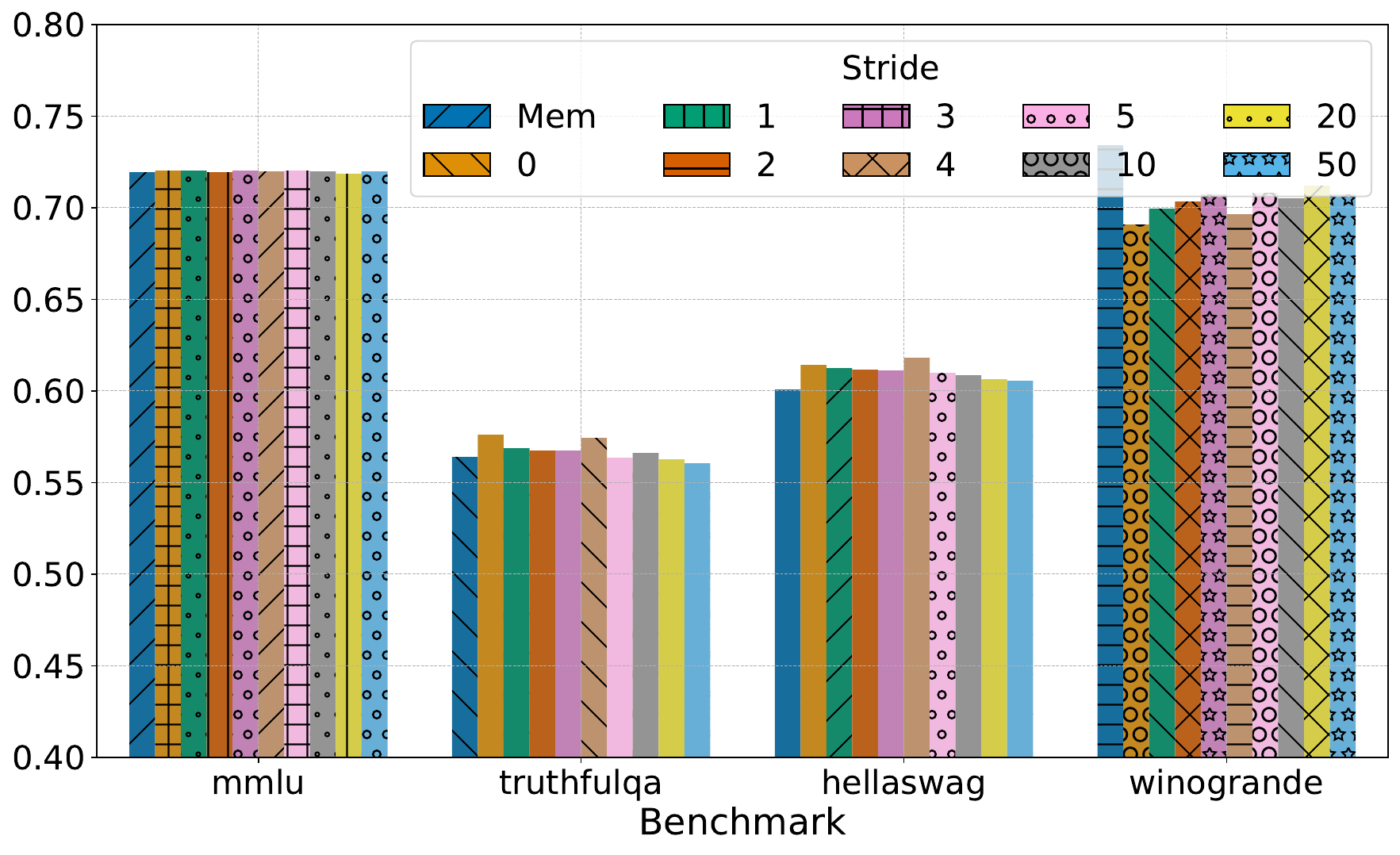}
\caption{Benchmark Qwen Synthetic}
\label{fig:strideBenchQwenSyn}
\end{subfigure}
\begin{subfigure}{0.49\textwidth}
\centering
\includegraphics[width=1\columnwidth]{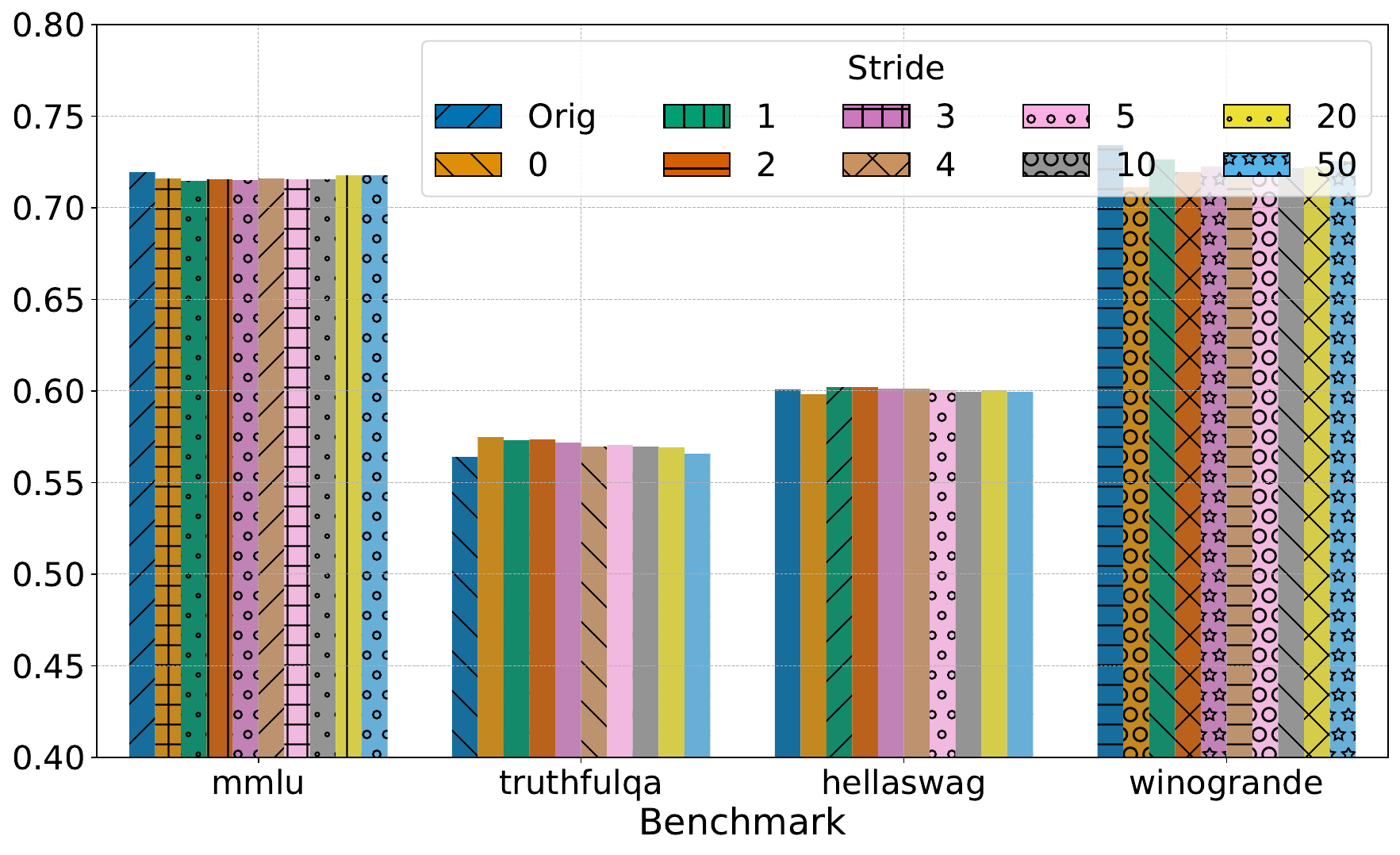}
\caption{Benchmark Qwen Organic}
\label{fig:strideBenchQwenOrg}
\end{subfigure}
\caption{Performance of \system{} with varying stride lengths. \autoref{fig:strideSyn} and \autoref{fig:strideOrg} illustrate the LCS on synthetic and organic data, respectively. \autoref{fig:strideBenchLlamaSyn} and \autoref{fig:strideBenchLlamaOrg} present benchmark results for synthetic and organic data on the Llama model, while \autoref{fig:strideBenchQwenSyn} and \autoref{fig:strideBenchQwenOrg} present results for the Qwen model.}
\label{fig:stride}
\end{figure*}

\subsection{Ablation Study}
We conduct extensive ablation studies to analyze the sensitivity of \system{} to various hyperparameters. Our analysis focuses on Llama and Qwen models using representative samples from both synthetic and organic datasets.

\mypara{Top-k}
We investigate the sensitivity of both forget ($\floss$) and maintain ($\mloss$) losses to the number of probability distributions considered, examining $k \in \{5, 10, 20\}$. Results shown in \autoref{fig:topK} for synthetic and organic data  demonstrate a strong stability across different $k$ values. This consistency extends to model utility metrics, suggesting that \system{} is robust to this hyperparameter choice at even modest coverage of the target probability distribution.

\mypara{Stride} 
The token stride parameter determines the granularity of unmemorization by controlling the spacing between modified tokens. We examine stride values spanning multiple orders of magnitude: {0,1, 2, 3, 4, 5, 10, 20, 50}. \autoref{fig:stride} presents the Longest Common Subsequence (LCS) results for synthetic and organic data.

Our results reveal the expected inverse relationship between stride length and unmemorization effectiveness, where decreasing stride length generally increases unmemorization. This is expected due to more token modifications that disrupt verbatim generation. However, it is important to note that the unmemorization remains significant even for larger strides. For example, a stride of 50—modifying only $\approx 2\%$ of tokens—\system{} reduces the LCS from over to maximum of 38 tokens. From these results, we observe that the unmemorization performance approximately saturates for strides up to 4, where we see nearly identical unmemorization performance. Therefore, we use a stride of 4 in our experiments.
The impact on model utility, illustrated in \autoref{fig:strideBenchLlamaSyn}  and \autoref{fig:strideBenchLlamaOrg} for Llama, and \autoref{fig:strideBenchQwenSyn} and \autoref{fig:strideBenchQwenOrg} for Qwen, demonstrates significant robustness of \system{} with a negligible effect on model utility. Contrary to initial intuition, even an aggressive stride setting of 0 (modifying every token) maintains model utility with minimal degradation. A closer examination reveals that this behavior is explained by the probability distribution of each token remaining similar with respect to the top 9 probabilities during unmemorization (after removing the target token). We hypothesize that maintaining this distribution over the top 9 probabilities is what helps the model preserve its utility. To further validate this hypothesis, we tested unmemorizing with a stride of 5 without applying the maintain loss $\mloss$. As anticipated, the model diverged and experienced a significant drop in utility.

\begin{figure}[!t]
\centering
\includegraphics[width=1\columnwidth]{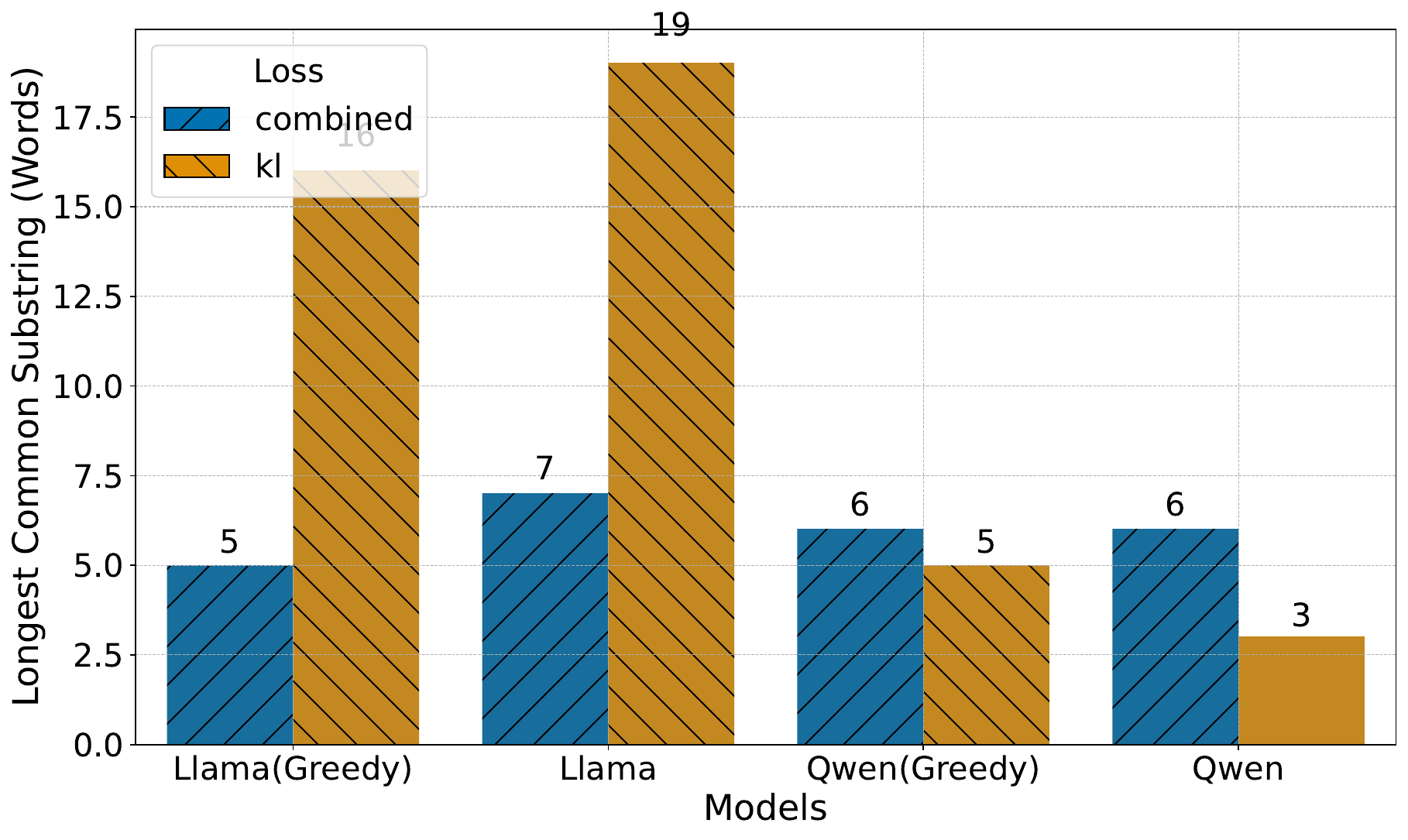}
\caption{Evaluation of using the $\fkl$ loss alone (KL) and the combined loss $\fkl + \ftarget$ (Combined).}

\label{fig:lossComb}
\end{figure}

\mypara{Influence Of Different Loss Functions} Next, we evaluate the effect of adding the probability loss ($\ftarget$) to the distribution loss ($\fkl$) when creating the forget loss $\floss$. To recap, the probability loss is primarily added to further reduce the probability of the target tokens to be unmemorized. We calculate the average probability of the target tokens, which shows that the probability is reduced by approximately one order of magnitude. We also benchmark models with the different loss constructions and find no significant effect. Furthermore, we plot the LCS in \autoref{fig:lossComb}, which shows that the combined loss achieves better performance than using only the distribution loss. Hence, we use the combined loss for our main settings.

\begin{figure*}[!t]
\centering
\begin{subfigure}{0.49\textwidth}
\centering
\includegraphics[width=1\columnwidth]{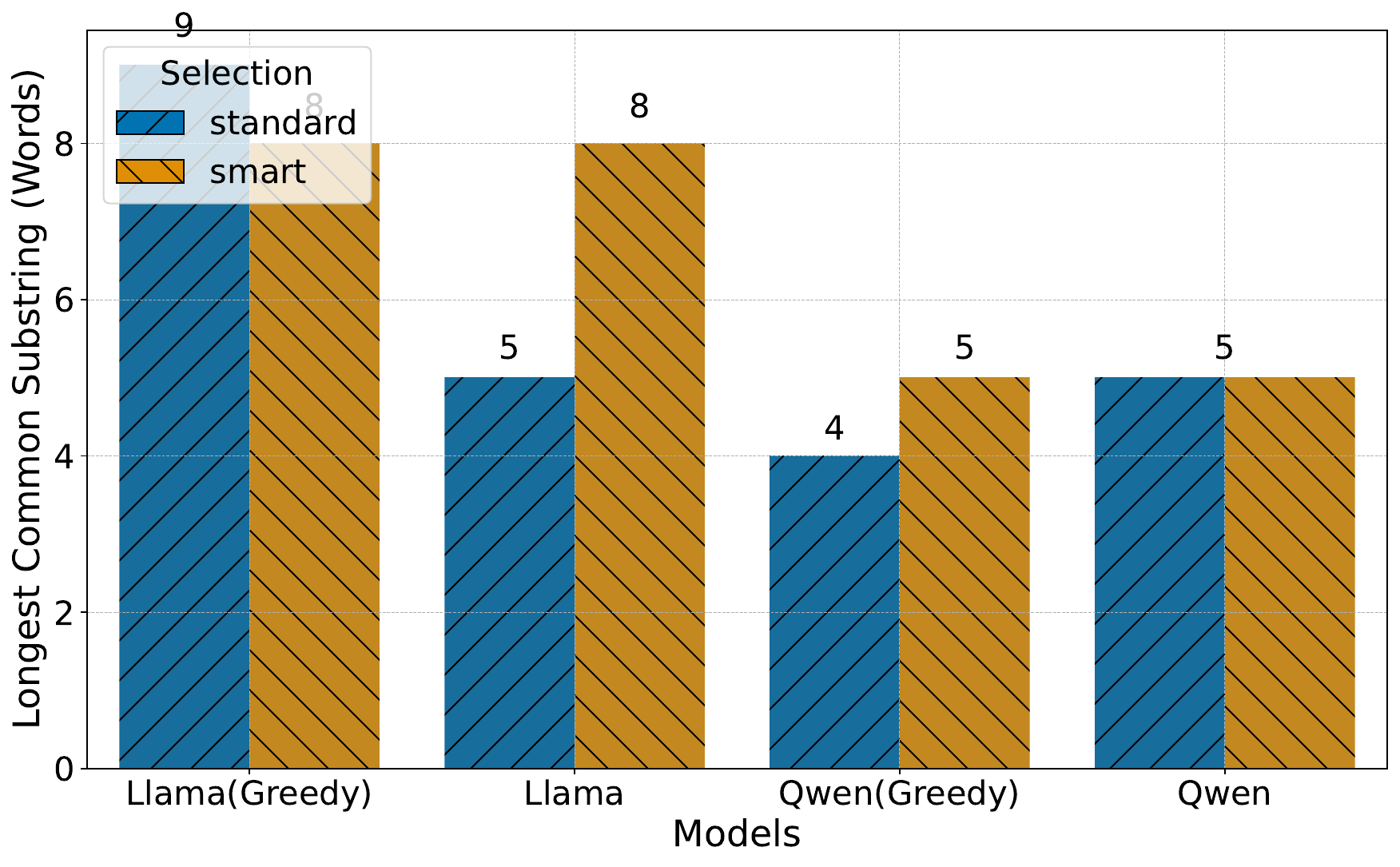}
\caption{
Synthetic (LCS $\downarrow$).}
\label{fig:candSelect_synthetic}
\end{subfigure}
\begin{subfigure}{0.49\textwidth}
\centering
\includegraphics[width=1\columnwidth]{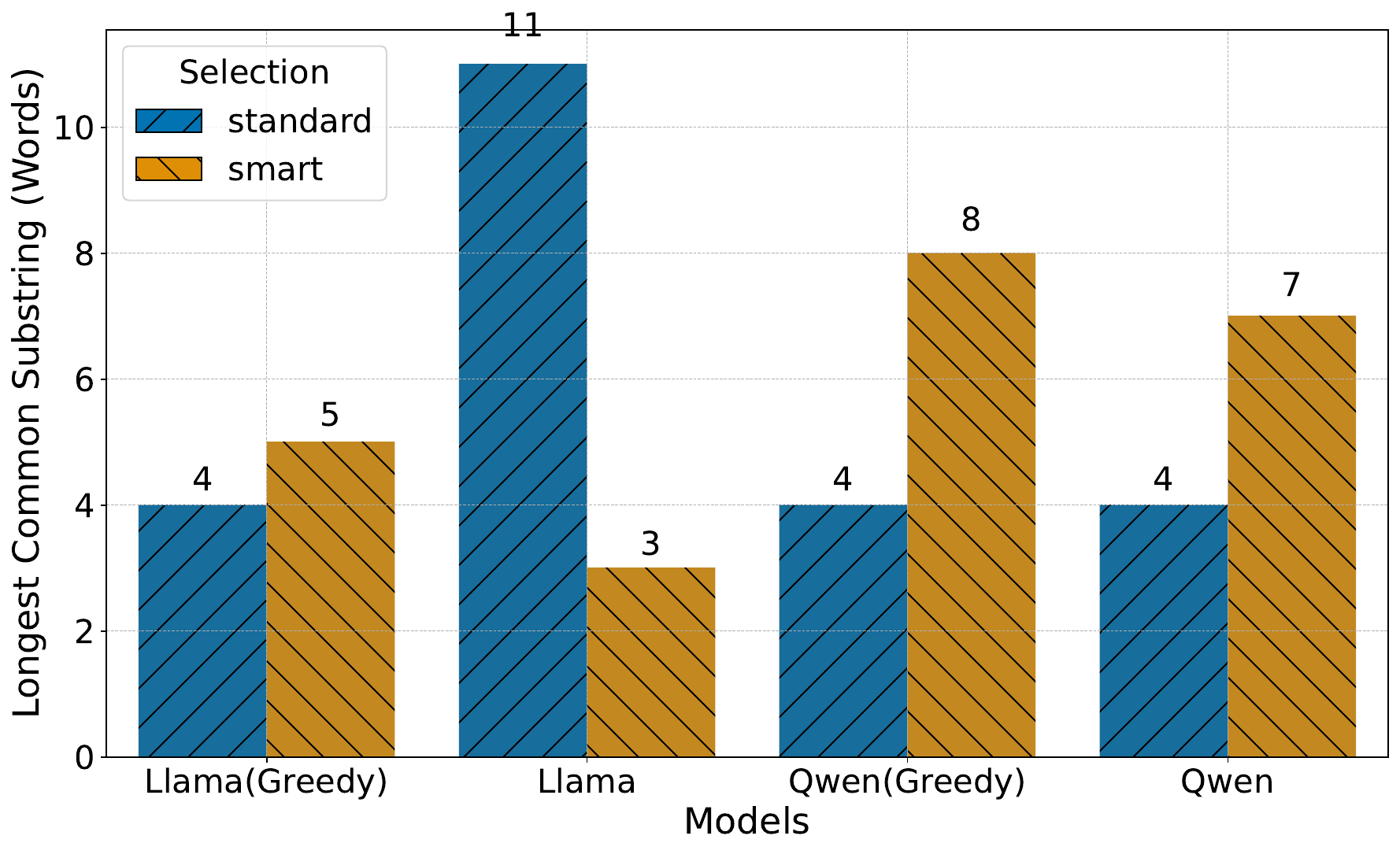}
\caption{
Organic (LCS $\downarrow$).}
\label{fig:candSelect_org}
\end{subfigure}
\caption{The effect of using the candidate selection techniques mentioned in \autoref{sec:meth} (Smart) versus using normal selection, i.e., no filtering of candidates (Standard).}

\label{fig:candSelect}
\end{figure*}

\mypara{Candidate Token Selection Strategy} Our final ablation examines the token candidate selection strategy for the forget loss ($\floss$). We compare two approaches: (1) direct selection from top-k probabilities, and (2) selection ensuring matching capitalization and leading space patterns for the $k$ alternate tokens (as mentioned in \autoref{sec:forgetLoss}). We present the results in \autoref{fig:candSelect} for both organic and synthetic datasets. As the figure shows, as expected both techniques have a similar performance with respect to unmemorization. However, benchmarking both techniques shows an advantage for using the smart selection of candidates, as it better preserves the model's behavior. This was especially monitored when using the MT-Bench in the CoTaEval evaluation; using the standard candidates resulted in a MT-Bench score of only 3.7 versus 4.67 for the smart selection, which shows that smart candidate selection can better help maintain utility.
These ablation studies collectively demonstrate that \system{} maintains robust performance across a wide range of hyperparameter configurations. The method exhibits particular stability in probability distribution coverage and candidate selection, while offering better unmemorization performance for smaller stride lengths without a significant loss in utility. We believe these results highlight the strong performance and stability of \system{} without the need to rely on precise hyperparameter tuning.

\section{Discussion and Limitations}
\label{sec:limitations}
Our work introduces \system{} as a targeted solution for preventing verbatim reproduction in LLMs, addressing a specific but crucial aspect of the broader AI copyright landscape. While our results demonstrate significant success in preventing exact sequence reproduction, it is important to clarify both the strengths and limitations of our approach within the larger scope of AI intellectual property protection.

\subsection{Limitations} 
Despite \system{}'s effectiveness in preventing verbatim generation, it is important to acknowledge a --by design-- limitation: while \system{} successfully prevents exact sequence reproduction, it does not address the broader challenge of information extraction from language models. A determined adversary could potentially extract information about the unmemorized content through careful prompt engineering, such as using a series of yes/no questions or constructing prompts that elicit semantic information without requiring exact reproduction. This limitation is inherent to our focused approach on verbatim generation and reflects the fundamental trade-off between maintaining model utility and completely removing information. Our method's effectiveness relies on the assumption that normal usage patterns primarily involve direct generation rather than adversarial probing. While this assumption aligns with typical use cases and current copyright frameworks that focus on expression rather than ideas, it does not satisfy a typical complete, i.e., both concept and verbatim, unlearning requirements.

\subsection{Broader Implications}
The success of \system{} in selectively mitigating verbatim reproduction while preserving semantic understanding further explore the area of knowledge representation in large language models. Our results indicate that surface-level text patterns can be modified effectively without significantly impacting semantic understanding, reinforcing the distinction between memorized sequences and learned concepts.

Furthermore, \system{} offers an efficient solution for model providers who need to address copyright concerns without resorting to complete retraining. By significantly reducing exact sequence matches while maintaining performance, \system{} aligns with emerging standards and ongoing discussions about copyright content protection.

\subsection{Challenges in Proving Training Data Usage}

Beyond copyright considerations, our findings also have direct implications for research on determining whether a model has been trained on specific data. Traditional membership inference attacks alone are insufficient for proving such claims, as recent work suggests that true verification requires data reconstruction, often using special canary data \cite{ZDKT24}. However, these proofs typically depend on verbatim reproduction—something that \system{} effectively prevents. As a result, once a model has been unmemorized using techniques like \system{}, existing approaches to training data verification may become less effective.  We anticipate that this challenge will spur new research directions focused on proving training data usage through more indirect methods
\section{Conclusion}
\label{sec:conc}
We introduced \system{}, a novel post-training approach to prevent verbatim text reproduction in large language models by modifying token distributions. Our results across multiple architectures show that precise control over memorized content can be achieved without compromising model utility. The success of token-level interventions in preventing exact sequence generation while preserving semantic understanding suggests potential for new techniques in targeted knowledge modification in neural networks, especially for selective removal of specific patterns while maintaining underlying learned representations.

\bibliography{ref}

\bibliographystyle{IEEEtran}
\end{document}